\documentclass[a4paper,12pt,times,numbered,print,oneside,index,custommargin]{PhDThesisPSnPDF}

\input{Preamble/preamble}

\title{Deep Learning-Based Classification of Hyperkinetic Movement Disorders in Children}

\subtitle{MRes Individual Research Project}

\author{Nandika Ramamurthy}

\dept{School of Biomedical Engineering and Imaging Sciences}

\university{King's College London}
\crest{\includegraphics[width=0.4\textwidth]{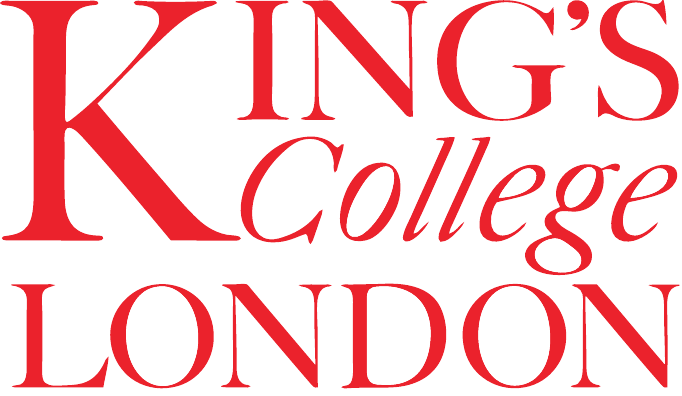}}

\supervisor{Dr Rachel Sparks\newline
Dr Daniel Lumsden}

\degreetitle{Masters of Research in Healthcare Technologies}

\subject{LaTeX} 

\ifdefineAbstract
\fi

\begin{document}

\mainmatter

\maketitle


\begin{declaration}

I hereby declare that except where specific reference is made to the work of 
others, the contents of this dissertation are original and have not been 
submitted in whole or in part for consideration for any other degree or 
qualification in this, or any other university. This dissertation is my own 
work and contains nothing which is the outcome of work done in collaboration 
with others, except as specified in the text and Acknowledgements. This 
dissertation contains fewer than 65,000 words including appendices, 
bibliography, footnotes, tables and equations and has fewer than 150 figures.


\end{declaration}


\begin{acknowledgements}

I am deeply grateful to my supervisor, Dr. Rachel Sparks, whose guidance and support have been pivotal in shaping this thesis. Her thoughtful feedback and guidance have enriched my work and inspired me to push my boundaries. I am equally thankful to Dr. Daniel Lumsden for his expert advice on the clinical aspects of this project, which was essential to its success.
My heartfelt appreciation goes to my friends and family for their constant encouragement. A special note of gratitude is owed to my partner, Thanis Murugathas, whose steadfast support and constant inspiration have made all the difference during the past few months.

\end{acknowledgements}

\begin{abstract}
\textbf{Introduction}\\
Hyperkinetic movement disorders (HMDs) in children mainly comprise dystonia (abnormal twisting) and chorea (irregular, random movements). HMDs present significant diagnostic challenges due to their complex and often overlapping clinical manifestations. Prevalence of these conditions ranges from 2 to 50 per million and 5 to 10 per 100,000 respectively. These patients typically experience diagnostic delays averaging 4.75 to 7.83 years. Traditional diagnostic methods rely on clinical history and physical examinations by an expert, as specialised tests do not improve detection due to their complex pathophysiology. To address challenges in diagnosis, this study develops a neural network model to differentiate between dystonia and chorea from video recordings of patients performing a specific task.\\

\noindent\textbf{Methods}\\
The model utilizes joint position data extracted from videos of paediatric patients performing motor tasks. The model architecture integrates a Graph Convolutional Network (GCN), to capture spatial relationships, and Long Short-Term Memory (LSTM), to represent temporal dynamics. Attention mechanisms were incorporated to refine feature extraction and improve the model’s interpretability. A dataset of 50 videos (31 chorea-predominant, 19 dystonia-predominant) was used to train and validate the model. All data was collected under regulatory approval by Guy’s and St Thomas’ NHS Foundation Trust (GSTT) Electronic Records Research Interface (GERRI) (REC\# 20EM/0112). \\

\noindent\textbf{Results and Discussion\\
}The model achieved an accuracy of 85\%, and an F1 score of 81\%. The model demonstrated high sensitivity (81\%) and specificity (88\%) at the optimal frame rate of 15 frames per second (fps). Attention maps provided valuable insights into the model's decision-making, highlighting the model’s correct identification of involuntary movement patterns, while model misclassifications are often linked to occluded body parts or subtle movement variations. This study highlights the potential of deep learning to improve the accuracy and efficiency of HMD diagnosis and ultimately could contribute to more reliable, interpretable clinical tools for paediatric HMDs.
\end{abstract}


\tableofcontents

\listoffigures

\listoftables




\chapter{Introduction}  

\ifpdf
    \graphicspath{{Chapter1/Figs/Raster/}{Chapter1/Figs/PDF/}{Chapter1/Figs/}}
\else
    \graphicspath{{Chapter1/Figs/Vector/}{Chapter1/Figs/}}
\fi

\section{Clinical Problem} 
\subsection{Types and Classification of HMDs} 
Paediatric movement disorders are broadly categorised into hypokinetic and hyperkinetic movement disorders. Hyperkinetic movement disorders (HMDs) are characterized by excessive or exaggerated movements (Stouwe et al., 2021). Hypokinetic movement disorders involve reduced voluntary movement (Kruer, 2015). The classification of paediatric movement disorders has primarily focused on HMDs as there are fewer types of hypokinetic movement disorders. Accurate and precise definitions of each type of HMD are essential for diagnosis, therapeutic decision-making, and effective communication between healthcare professionals.\\

The prevalence of HMDs in the paediatric population is not fully established due to the lack of comprehensive studies and the challenges associated with diagnosing these disorders (Kruer, 2015). There are specific estimates for only certain types of movement disorders. For instance, dystonia is observed in 2 to 50 cases per million individuals under the age of 26 (Fernández-Alvarez and Nardocci, 2012). It is also commonly seen in dyskinetic cerebral palsy (CP), occurring 2–3 per 1000 live births (Lin, 2011). The prevalence of chorea is less clear due to limited studies, although conditions like Huntington’s disease, which is one of several causes of chorea, is found in 5-10 cases per 100,000 individuals, and benign hereditary chorea occurs in approximately 1 per half a million individuals (Merical and Sánchez-Manso, 2024). The cited prevalences likely underestimate the true incidence in the population, due to the misdiagnosis of atypical presentations and insidious onset of symptoms (Fernández-Alvarez and Nardocci, 2012). Given the prevalence and complexity of HMDs, accurate classification and characterization is crucial to advancing research in their diagnosis and treatment. The following section will summarize the clinical presentation of two prominent HMDs—dystonia and chorea—and highlight the challenges in distinguishing between them.\\

 \begin{figure}[h!]
    \centering
    \includegraphics[width=0.7\textwidth]{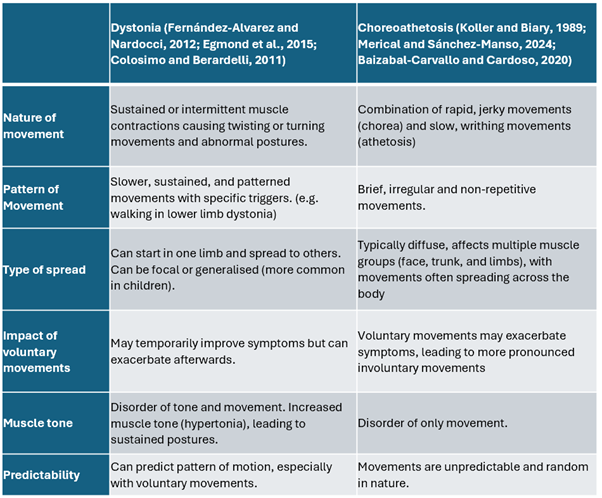}
    \caption{Key features that distinguish dystonia from choreoathetosis based on their clinical presentation.}
    \label{fig:image-label}
\end{figure}

Dystonia is characterized by abnormal twisting or posturing movements, which can affect either a single part of the body (focal) or multiple regions (generalized) (Fernández-Alvarez and Nardocci, 2012). Diagnosing dystonia can be challenging, particularly when it presents with spasticity in conditions such as cerebral palsy, where the resulting “mixed tone” mimics patterns of multiple HMDs (Gorodetsky and Fasano, 2022). Dystonia can present alongside myoclonus, with more pronounced involvement of the upper body (Sanders et al., 2024). Developmentally normal children may be misdiagnosed due to “overflow movements” caused by an immature nervous system (Gorodetsky and Fasano, 2022).\\

In contrast, chorea presents as irregular, random, and brief movements that flow from one part of the body to another, often giving the appearance of restlessness. These continuous movements are generally considered non-suppressible, although one study found that 50\% of patients could temporarily suppress their chorea (Koller and Biary, 1989). Additionally, ballismus is a type of chorea that affects the proximal limb (i.e. shoulder). It is a high-amplitude, forceful flinging motion of proximal joints (Nikkhah et al., 2019). Associated forms with chorea involve athetosis, a slower, smooth, and writhing movement typically affecting the distal parts of the limbs. Athetosis is frequently observed in dyskinetic CP and often co-occurs with dystonia. The term “choreoathetosis” is used by clinicians to describe movements that exhibit characteristics of both chorea and athetosis, as they often occur in combination (Yilmaz and Mink, 2020). \\

Understanding the pathophysiology of different HMDs is essential for developing targeted treatments and predicting disease progression. Knowledge of the underlying disease processes can aid in distinguishing between different disorders to ensure accurate classification.\\

  \begin{figure}[h!]
    \centering
    \includegraphics[width=0.7\textwidth]{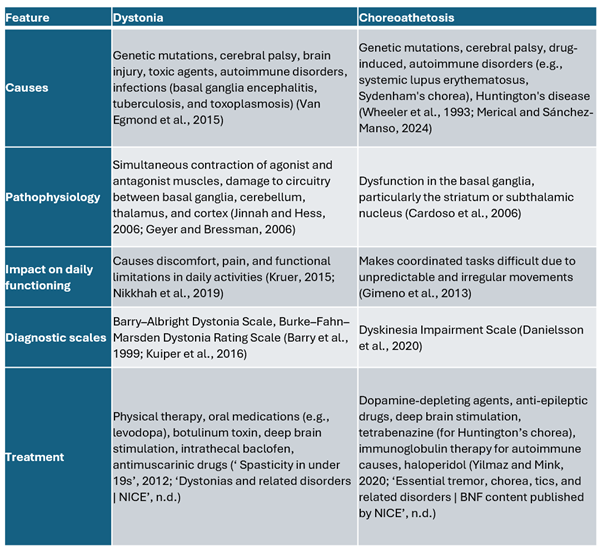}
    \caption{Summary of the pathophysiology and treatment of dystonia and chorea.}
    \label{fig:image-label}
\end{figure}

\subsection{Clinical Diagnosis and Treatment}

Diagnosis of HMDs relies on clinical history and a detailed physical examination (Mink and Sanger, 2017). Quantifying the severity of the disorder is also important to understand the progression of the disease and the effect of therapeutic interventions. Scales for assessing HMDs exist, but they tend to focus on one type of movement disorder and vary widely in their diagnostic properties, availability, and administration effort (Pietracupa et al., 2015). These systems may also need to be altered for the paediatric population by replacing more advanced assessment tasks. \\

The Barry–Albright Dystonia Scale (BADS) and Burke–Fahn–Marsden Dystonia Rating Scale (BFMDRS) are the most widely reported scales for childhood dystonia (Barry et al., 1999; Kuiper et al., 2016). BADS is widely used as it has a low administration effort, assessing 8 specific body regions for dystonia from a scale of 0 to 4 (Barry et al., 1999) but does not characterise choreoathetosis. The BFMDRS offers a detailed evaluation of dystonia, including both motor symptoms and the disability they cause in daily activities (Stewart et al., 2017). \\ 

Currently, there are no specific diagnostic tests to distinguish dystonia from chorea, as clinicians must rely on recognising the nature of abnormal movements to differentiate between them. Existing scales measure the extent of dystonia or chorea but are limited in their ability to accurately diagnose and assess severity. To improve diagnosis, reduce variability among observers and better target treatments, more objective methods are needed. Further research is required to evaluate the effectiveness of clinical scales in characterizing HMDs and to ensure these scales are objective, time-efficient, and applicable across the various types.\\

\section{State-of-the-art} 
\subsection{Neural Networks for Capturing Spatiotemporal Dependencies}

Recognizing and diagnosing movement abnormalities can use features broadly classified into spatial (body part arrangement) or temporal dependencies (motion over time). Spatial features can detect abnormal postures or movement amplitudes, while temporal features reveal irregular patterns or movement frequency.\\

To capture spatial connections between joints, the human skeleton can be represented as a graph for input into Graph Convolutional Networks (GCNs). GCNs are a type of CNN designed to work with graph-structured data, making them ideal for learning spatial relationships in the graph (Kipf and Welling, 2017). In this context, skeleton joints are treated as graph nodes, and their connections as edges (Wu et al., 2021). GCNs transform these into feature vectors, like how CNNs process grid-like data (Defferrard et al., 2016). After applying a convolution, a weighted average is used to aggregate the features. Despite their effectiveness in capturing joint relationships, GCNs require normalization due to sensitivity to input scale for stable training (Khemani et al., 2024).\\

\[
H^{(l+1)} = \sigma \left( \tilde{D}^{-1/2} \tilde{A} 
\tilde{D}^{-1/2} H^{(l)} W^{(l)} \right)
\]where
\[
\tilde{A} = A + I_N
\]
and
\[
\tilde{D}_{ii} = \sum_j \tilde{A}_{ij}
\]
\textbf{Equation 1:} This summarises the GCN mechanism, where \( H \) is a matrix of node features at the \( l \)-th layer, where features are aggregated through an adjacency matrix \( A \) and normalised by the degree of the nodes \( D^{\sim -1/2} \). This is transformed by multiplying with the weight matrix \( W \) and an activation function \( \sigma \) is applied to learn complex patterns. (Wu et al., 2019)

\[
H^{(l+1)} = \sigma \left( \tilde{D}^{-1/2} \tilde{A} \tilde{D}^{-1/2} H^{(l)} W^{(l)} \right)
\]

To capture temporal features, Recurrent Neural Networks (RNNs) can be used to model sequential data by using previous outputs \( h_{t-1} \), encoded as hidden states, along with the current input \( x_t \) to produce a predicted output \( h_t \) for the time point. These hidden states help the network learn temporal dependencies. However, RNNs can struggle with vanishing or exploding gradients during training, limiting their effectiveness on long sequences (Pascanu et al., 2013). To address this limitation, Long-Short Term Memory (LSTM) was developed with gated memory units to retain important information and discard irrelevant data (Hochreiter and Schmidhuber, 1997).\\ 

 \begin{figure}[!ht]
    \centering
    \includegraphics[width=0.7\textwidth]{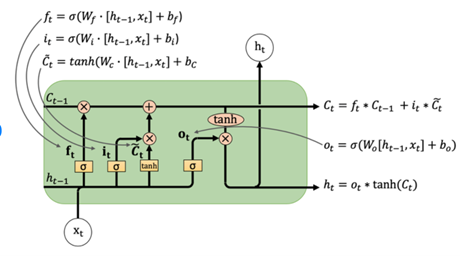}
    \caption{An LSTM Cell. This shows how a new cell state \( C_t \) can be created each time by combining a retained old state \( C_{t-1} \) and a candidate new state \( \tilde{C}_t \). The basic LSTM unit has three gates: the input gate \( i_t \), output gate \( o_t \), and forget gate \( f_t \), hence capturing long-term dependencies in data more effectively (Zhao et al., 2023).}
    \label{fig:image-label}
\end{figure}

Attention mechanisms are another valuable network component that can enhance feature representation and adaptability to varied inputs. They enable the model to focus on key parts of the input sequence for better output prediction. The mechanism transforms each input element into three vectors: query (the element seeking information), key (elements in the sequence used for comparison), and value (information corresponding to each key). Attention weights are calculated by the dot product of the query and key, followed by SoftMax. These weights help capture dependencies within the data. In self-attention, the query, key, and value are the same, while multi-head attention uses multiple sets of these vectors for the final output. (Vaswani et al., 2023; Soydaner, 2022)\\

\[
\text{Attention}(Q, K, V) = \text{Softmax}\left( \frac{Q K^T}{\sqrt{d_k}} \right) V
\]
\textbf{Equation 2:} This equation represents the attention mechanism in terms of the query \( Q \), key \( K \), and value \( V \). The dimension of key vectors \( d_k \) is used to form a scaled dot product to keep the gradients of the network in a manageable range. (Vaswani et al., 2023)\\

The neural networks mentioned above have been employed to capture spatial and temporal information from human poses for action recognition (Qin et al., 2022) and movement prediction (Shu et al., 2022). For example, a study proposed the Skeleton-Joint Co-Attention Recurrent Neural Network (SC-RNN) to predict future motion of a skeleton representation of human movement (Shu et al., 2022). It maps the spatial and temporal coherence of joints onto a skeleton-joint co-attention feature map, which is then used with a Skeleton-Joint Co-Attention (SCA) mechanism to allow more informative features to predict motion (Shu et al., 2022). The model measured the coherence of joint motion with neighbouring joints and the coherence between observed motion and future motion of the entire skeleton. The model demonstrated a lower Mean Absolute Error (MAE) compared to other state-of-the-art models. However, it showed increased errors in longer sequences and sometimes predicted unnatural motions.\\

A study introduced the ST-GCN (Spatial Temporal GCN) for skeleton-based action recognition, which captures spatial connections within the skeleton at each frame and temporal connections across frames (Yan et al., 2018). While spatial relations of GCNs have been described in previous literature, this method uses GCNs to capture temporal connections too. The model used hierarchical and localised representations of the joints and their trajectories by restricting the model to the local regions of the graph. A hierarchical representation involving multiple GCN layers captured finer details at lower layers and more abstract features at higher layers to reduce the overall complexity of the model.\\

Hierarchical network architectures are explored in a study proposing a Hierarchical Spatial Reasoning and Temporal Stack Learning network (HSR-TSL), which uses GNN intra-parts and inter-parts networks to capture spatial relations within each body part and between body parts respectively (Si et al., 2020). Temporal dynamics were captured by dividing videos into short clips and processing each clip using an LSTM. Although the spatial and temporal GCNs did not use co-occurrence information, the model achieves high accuracy compared to other state-of-the-art models (87.7\% in cross-subject settings).\\

Another study presented a self-attention mechanism to improve GCN performance for skeleton-based action recognition, achieving 90.5\% accuracy (Qin et al., 2022). The model represented connected the input data using a directed graph to improve the model performance. Multi-head attention was explored in a study introducing an action transformer (AcT) that utilised minimal architectural priors in the network, allowing a less computationally intensive network with improved generalisation capability (Mazzia et al., 2022).\\

\subsection{Models for Joint and Motion Tracking}
Many of the previously discussed models require an algorithm to detect joints as a preprocessing step, also known as human pose estimation. It is defined as the localization of human joints or predefined landmarks in images or videos to estimate the body’s configuration (Chen et al., 2020). Whole-body pose estimation involves localising keypoints of the body, face, hand, and foot simultaneously, and is a challenging problem tackled using deep learning-based approaches (Jin et al., 2020).\\

Pose estimation algorithms follow two main approaches: top-down and bottom-up. In the top-down approach, models first detect a person often by using a bounding box after which individual body parts are identified. The bottom-up approach detects individual body parts first and then groups them to form a pose for one person. MoveNet is a notable top-down model, while OpenPose, DeepLabCut, and PoseNet are popular bottom-up models. Bottom-up methods are more efficient as they do not need to rerun for each person in a frame (Osokin, 2018). However, they can suffer from scale variances when multiple people are in the frame (Jin et al., 2020).\\

OpenPose is a deep learning model that provides 2D keypoints for the whole body. It first detects body and foot keypoints and then uses these to locate hand and face keypoints (Cao et al., 2019). OpenPose is highly accessible, running on various operating systems and hardware. It uses heatmap estimation at each stage to identify keypoints, enhancing robustness to noise while preserving gradient information (Luo et al., 2021). OpenPose can distinguish front and back unlike other pose estimation methods, thus reducing error in tracking joint positions (W. Kim et al., 2021). It uses part affinity fields to encode the location of the limbs as a vector magnitude and the orientation of the limbs as a vector direction (Jin et al., 2020). Optimised versions use dilated convolutions to increase the receptive field of the network by introducing gaps between kernel elements and reusing trained weights (Osokin, 2018).\\

DeepLabCut is a pose estimation software that uses a variant of ResNet. It applies transfer learning to estimate 14 keypoints, adapting the model to new datasets using pre-trained weights (Nath et al., 2019). PoseNet employs pre-trained networks (ResNet-50 and MobileNetV1) to generate heatmaps for joint prediction, refining keypoints with offset vectors (Kendall et al., 2016). MoveNet uses a feature pyramid network with residual connections to localize 17 keypoints of the body (Joshi and Joshi, 2021). In a comparative study between OpenPose, MoveNet and DeepLabCut, OpenPose and MoveNet produced the lowest error in measuring hip kinematics, and OpenPose produced the lowest error in measuring knee kinematic errors (Washabaugh et al., 2022).\\

OpenPose and other platforms have been trained and validated on able-bodied individuals, therefore, it is difficult to predict model performance in movement disorders where postural differences, impairments and physical deformities may lead to errors in pose estimation (Washabaugh et al., 2022). One study has demonstrated that OpenPose is effective in measuring joint angles in populations with postural abnormalities, even with occlusions and non-frontal views (W. Kim et al., 2021). However, in another study, OpenPose’s performance was 30\% lower than other methods when incorrect camera position and self-occlusion were introduced (Chung et al., 2022). Therefore, these conditions need to be optimised to improve pose estimation.\\

\subsection{Models for Diagnosis of Movement Disorders}
Currently, no computational models can distinguish between different types of HMDs in patient care (Méneret et al., 2021). However, similar approaches for other medical conditions that affect movement may serve as a foundation for powerful HMD diagnostic tools.\\

A proof-of-concept study employed a Random Forest algorithm to detect dystonia in individuals with dyskinetic CP (Haberfehlner et al., 2023). Using three limb length measurements from video frames, the model was trained with clinician-labelled data using the Dyskinesia Impairment Scale. This method demonstrated effective performance due to greater control over feature selection. However, the video recordings were labelled by three independent raters which resulted in moderate inter-rater variability, which complicates the establishment of a reliable “ground truth,” hence questioning the trustworthiness of the labels. Additionally, the model did not analyse left and right limbs separately, which may have overlooked asymmetrical movement patterns.\\

A follow-up study aimed to classify dystonia and choreoathetosis of the distal leg using short video sequences (Haberfehlner et al., 2024). It employed Hierarchical Vote Collective of Transformation-based Ensembles (HIVE-COTE 2.0), which are four time series-based classification algorithms. The study uses new pre-processing strategies including gap-filling to predict missing values in a dataset using multivariate iterative imputation, noise filtering, spatial normalization to correct camera angles, and temporal normalization to standardize video lengths. The study shows an accuracy of 81\% for dystonia but only 53\% for choreoathetosis, due to data scarcity for the latter class. Limitations of the study included difficulty in clinically validating automatically generated features and potential loss of critical information due to fixed video lengths.\\

ResNet-50-based neural network (He et al., 2015a) has been presented to diagnose dystonia using videos, achieving 93\% sensitivity and 93\% specificity (Miao et al., 2020). This approach uses clinically relevant features including inter-knee distance variance, foot angle variance and median difference of foot angle between limbs. These variables were calculated offline once the parts were identified by DeepLabCut. Similar work used a pre-trained ResNet50 model to classify head movement states (e.g., ‘face forward’, ‘tilt left’) (Peach et al., 2023). It was trained and validated using videos from both healthy controls and patients with dystonia, achieving an accuracy of 84\%.\\

Random Forest and 3D ResNet-18 (Hara et al., 2017) were evaluated on their ability to detect tics from facial features on video (Brügge et al., 2023). Bounding boxes were used to crop the videos to only the face regions and gradient-weighted class activation mapping (GradCAM) was used to generate heatmaps to visualise the facial regions that are important for the model prediction (Selvaraju et al., 2016). It improved the interpretability of the network’s classifications and highlighted misclassification such as classifying instances of blinking as tics. Random Forest and the 3D ResNet-18 achieved similar accuracies (88\%) and F1 scores (80-82\%).\\

In conclusion, while the field of automated deep learning models for distinguishing between different types of HMDs in patient care is still in its infancy, the exploration of similar diagnostic approaches for other medical conditions that affect movement is paving the way for future advancements.\\

\section{Socioeconomic and Ethical Considerations}  
\label{section1.3}
For deep learning models to effectively classify movement disorders, they must be inclusive across diverse populations to ensure equal representation and patient-centred care (Noseworthy et al., 2020). Incorporating varied demographic data in model training helps mitigate potential biases, which often arise from limited datasets that historically favoured able-bodied male subjects and excluded disabled populations (‘AI Is Being Built on Dated, Flawed Motion-Capture Data - IEEE Spectrum’, n.d.). A systematic review of AI diversity and inclusion shows the growing emphasis on mitigating bias, ensuring fairness, and enhancing transparency in models (Shams et al., 2023), which aligns with the ethical principles of equality and non-discrimination in medical practice (De Micco and Scendoni, 2024). This approach is essential to prevent AI from perpetuating or exacerbating societal biases in healthcare.\\

Socioeconomic factors significantly impact timely access to diagnosis and treatment for movement disorders. Specialists report that conditions like dystonia are often underestimated (Albanese et al., 2006), which leads to substantial diagnostic delays. For example, primary/primary-plus dystonia patients typically wait 4.75 years, while children with secondary dystonia experience delays of an average of 7.83 years. Delayed referrals result in delayed treatment and lowered quality of life while a definitive diagnosis is established. The study also noted referral bias, suggesting the study sample may not represent all children with dystonia but rather those severe enough to be referred to specialised services (Lin et al., 2014). Additionally, a study of dopa-responsive dystonia patients found that diagnostic delays have increased from 9 to 15 years since 1994 despite advances in genetic testing. However, selection and reporting biases may partially explain these results (Tadic et al., 2012). \\

Genetic testing for rare inherited diseases such as primary dystonia can increase pathway costs by £939 per patient in the UK (Payne et al., 2018). Accessibility to genetic testing varies widely with lower rates outside Europe and North America, thus creating significant disparities. Financial barriers in countries that rely on private healthcare also contribute to these inequalities (Gatto et al., 2021). Therefore, improved recognition of clinical presentations in movement disorders can lead to more targeted and cost-effective genetic screening.\\

Implementing an effective deep-learning model for diagnosing movement disorders could reduce delays in diagnosis and ensure accurate assessments. This approach could also lower costs for patients and healthcare professionals, as consultant time for diagnosis could be reduced, thus improving resource allocation and patient outcomes. Ensuring the model is trained on consultant-led diagnoses and validated scales maintains clinical standards by ensuring standardization across clinics. Therefore, transparency in its mechanisms may ensure alignment with established medical practices.\\

\section{Aims}
This proof-of-concept project aims to develop a robust neural network capable of differentiating between dystonia and chorea by analysing joint position data extracted using OpenPose, a software tool for human pose estimation. The proposed network integrates a GCN extract features for the spatial relationships between joint positions, and a LSTM, to extract features for the temporal dynamics of joints over time.\\

The performance of the network is evaluated using the accuracy, sensitivity, and specificity of the final predicted diagnosis (chorea or dystonia). Moreover, to enhance the transparency and interpretability of the model, attention mechanisms are added to enable weights of the features in the model contributing toward the final prediction. This will help provide insights into how the model differentiates between dystonia and chorea.\\


\chapter{Methodology}

\ifpdf
    \graphicspath{{Chapter2/Figs/Raster/}{Chapter2/Figs/PDF/}{Chapter2/Figs/}}
\else
    \graphicspath{{Chapter2/Figs/Vector/}{Chapter2/Figs/}}
\fi

    \section{Data Collection and Pre-Processing}
\subsection{Patient Selection and Video Preparation}
This study used videos collected from the routine assessment of children and young people selected for deep brain stimulation (DBS) at the Evelina Children’s Hospital (ECH). ECH is a supra-regional centre, with patients with complex, medication-refractory HMDs being referred from across the UK. Video recordings comprise patients being asked to perform various clinical tasks assessed using the BFMDRS recorded at baseline (before surgery) and then yearly following surgery. In this work, baseline assessment recordings were selected for further analysis.\\

Regulatory approval was granted by the Guy’s and St Thomas’ NHS Foundation Trust (GSTT) Electronic Records Research Interface (GERRI) (REC\# 20EM/0112), under the supervision of the Children’s Neurosciences department (‘GSTT Electronic Records Research Interface (GERRI)’, n.d.) for the use of videos from the last 100 children who had undergone DBS before December 2023.\\

Video recordings of baseline motor function for patients diagnosed with dyskinetic cerebral palsy (CP), and dystonia-predominant and chorea-predominant genetic disorders were selected. Patients were divided into two classes: chorea (dyskinetic CP and GNAO1) as the negative class, and isolated genetic dystonia (DYT-TOR1A, DYT-KMT2B and DYT-THAP1) as the positive class. Clinical diagnoses of these two groups are “Chorea predominant” and “Dystonia predominant”, rather than pure chorea and dystonia, as patients with mixed presentation were included in the study to better reflect the patient population seen in neurology clinics.\\

Videos were selected of a single task involving patients holding their hands upwards and outstretched for a few seconds. This task was chosen due to its simplicity and minimal need for staff assistance, making it accessible for patients who might have difficulty performing more complex tasks, such as walking unaided. Additionally, this task was chosen to minimize self-occlusion, as seen in tasks like the finger-to-nose test, ensuring clear detection of joint position.\\

  \begin{figure}[h!]
    \centering
    \includegraphics[width=0.7\textwidth]{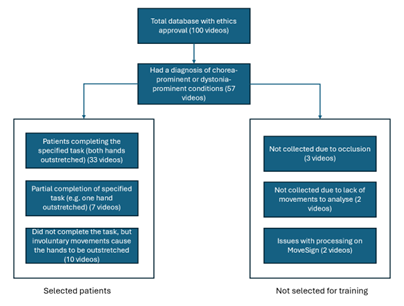}
    \caption{Video selection criteria. Only videos where patients successfully raised their hands for a minimum of three seconds were included in the final dataset to allow for temporal analysis.}
    \label{fig:image-label}
\end{figure}

Videos were screened to ensure the quality and relevance of the video data using the process depicted in Figure 2 including criteria to ensure that videos had optimal lighting conditions to maximize keypoint detection accuracy, a clear white background for contrast with the subject, and a stable, fixed camera position to avoid shaky footage. Selected videos depicted patients in a standardized sitting position and directly facing the camera, with all joints fully visible. The final dataset contained 50 high-quality videos for model training, with 31 videos of chorea-prominent presentations and 19 videos of dystonia-prominent presentations.\\

Videos were spatially cropped to remove any visual interference from staff. Videos were temporally trimmed to contain only frames related to the specific task. All video processing was performed using VLC (VideoLAN Client) (‘VideoLAN’, n.d.). Videos were resampled to have a frame rate of 25 frames per second (fps) to ensure uniformity in the temporal domain.\\

\subsection{Pose Extraction}
  \begin{figure}[h!]
    \centering
    \includegraphics[width=0.7\textwidth]{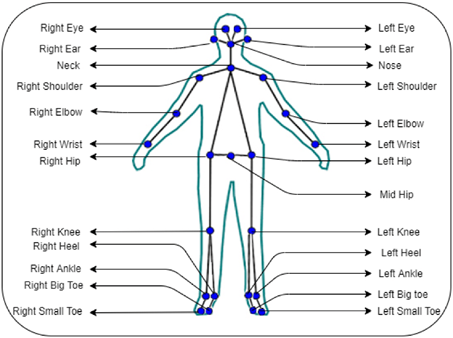}
    \caption{Skeleton keypoints extracted from videos using OpenPose (Shopon et al., 2021).}
    \label{fig:image-label}
\end{figure}

The MoveSign application was used, which is a lightweight wrapper to OpenPose with specific features to maximise tracking and keypoint detection. The OpenPose model “Body\_25” extracted 25 keypoints from each video as seen in Figure 3. Resolution-based normalization helped map the keypoint coordinate values between 0 and 1. 
To enhance the accuracy of keypoint detection, only keypoints detected with a confidence level greater than 5\% were considered. Non-Maximum Suppression (NMS) was employed within each frame to prevent multiple detections of the same keypoint retaining only the most confident detection, enabling precise identification of keypoints.\\

MoveSign enabled tracking to ensure the OpenPose model estimates the positions of keypoints even when occluded. Although these estimates are less precise, they maintain keypoint tracking continuity. MoveSign also implemented a smoothing algorithm to reduce jitter, thus ensuring stable detection of keypoints over time.\\

\subsection{Creating the Input Matrices}
The input spatiotemporal matrix was comprised of the x and y coordinates of 25 keypoints over 75 consecutive frames from the video (forming a matrix with the dimensions of batch size x 50 x 75). For the spatial network input, this matrix was divided into keypoints corresponding to the left arm, right arm, left leg, right leg, and torso following the hierarchical structure of human movement.\\

Keypoints were recorded to form a travelling sequence as shown in Figure 4 to better model relationships between body parts. This method of ordering is supported by research indicating this arrangement is more effective in capturing the dynamics of human movement (Shu et al., 2022).\\

   \begin{figure}[h!]
    \centering
    \includegraphics[width=0.7\textwidth]{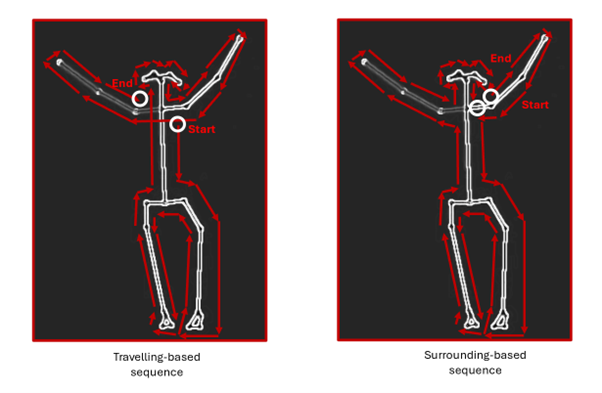}
    \caption{Two methods for keypoint ordering recreated from Shu et al., 2022. White circles indicate the start and end of each sequence. The travelling-based sequence was found (left) to have a lower mean angle error for predicting joint motion compared to the surrounding-based sequence (right).}
    \label{fig:image-label}
\end{figure}

\section{Network Architecture}
  
    \begin{figure}[h!]
    \centering
    \includegraphics[width=0.7\textwidth]{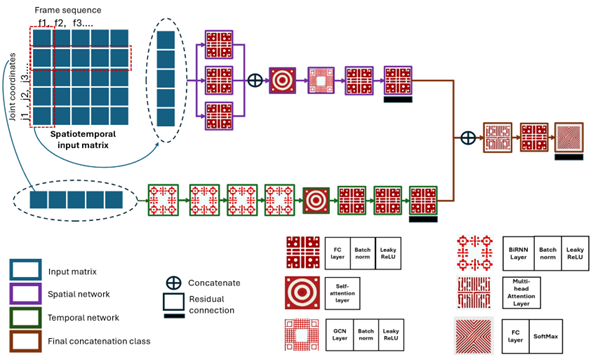}
    \caption{A block diagram of the network architecture. FC Layer = Fully Connected Layer, BiRNN = Bidirectional RNN.}
    \label{fig:image-label}
\end{figure}

     \begin{figure}[h!]
    \centering
    \includegraphics[width=0.7\textwidth]{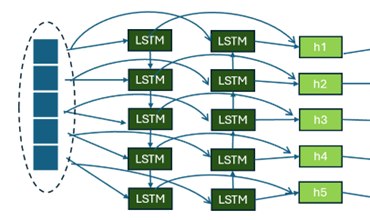}
    \caption{An illustration of a bidirectional LSTM layer, where h\_t represents the hidden states for time t, which is input into the next bidirectional LSTM layer.}
    \label{fig:image-label}
\end{figure}

      \begin{figure}[h!]
    \centering
    \includegraphics[width=0.7\textwidth]{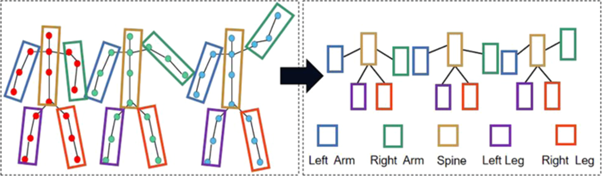}
    \caption{Illustration from Feng et al., 2022 showing part-level features in bounding boxes (left) and body-level features (right) with each part being a different colour. In this network, parts are encoded by FC layers to capture relationships between keypoints, and body-level features are encoded by a GCN to represent relationships between parts.}
    \label{fig:image-label}
\end{figure}

The network is designed with three components: a spatial network (top pathway in Figure 2.4), a temporal network (bottom pathway in Figure 2.4) and a combination of the two (right pathway in Figure 2.4) to generate a final combined classification.  This approach follows recent studies in the field (as discussed in Section 1.2.1), to allow the network to learn spatial features independently from temporal dynamics to locate the important body parts and video frames.\\

Every layer in the network is followed by the activation function Leaky ReLU due to its ability to handle negative and zero gradients effectively to overcome the vanishing gradient problem by allowing a small, non-zero gradient when the input is negative, where a\_i is a fixed parameter (typically 0.01) (Xu et al., 2015). Batch normalization is used to normalize the inputs to each layer, ensuring that the network converges quickly and effectively (Ioffe and Szegedy, 2015).\\

The spatial network models the spatial relationships between joints and captures the graph-like connections between different body parts within each video frame. This network uses three fully connected (FC) layers to encode the coordinates of each body part into a feature space. The FC layers share weights for symmetrical body parts (i.e. arms, legs) while maintaining distinct inputs and outputs, thus preserving unique characteristics for each side. This method enhances detecting similarities in motion patterns and improves generalization and efficiency (Si et al., 2020). The outputs of the FC layers represent relationships between the joints in each part.\\

After encoding each body part, outputs from the FC layers are combined and input to a self-attention layer, to give a higher weight to the most relevant body parts for the final classification decision. The data then passes through a GCN to capture relationships between body parts. The GCN allows the network to represent interactions between non-adjacent parts, such as arm-leg coordination. Self-connections were excluded to ensure the network learns only meaningful relationships between parts. Max aggregation was performed after the GCN to select the most representative features (Li et al., 2020). The GCN is then followed by two FC layers to learn high-level spatial relationships of the skeleton within each frame.\\

The temporal network models relationships between frames to recognize the motion of joints over time using four bidirectional LSTM layers to capture temporal dependencies. The bidirectional LSTM differentiates poses that vary in time such as waving versus raising a hand. A self-attention layer weights important frames and helps to refine the temporal features. Finally, three FC layers learn high-level temporal features across frames. A residual connection is added to the final two FC layers to aid gradient flow and speed up convergence during training (Philipp et al., 2018).\\

The spatial and temporal network output features are combined using a multi-head attention layer, enabling the network to learn which features from both networks contribute most to the final classification and a residual connection to aid gradient flow. The integrated features then pass through two FC layers, condensing features, followed by a SoftMax layer to determine the logistics of the video clip corresponding to the final binary classification task (chorea versus dystonia).\\

\section{Loss Function}
In the context of classification involving imbalanced datasets traditional loss functions like Cross-Entropy may fail to achieve optimal performance as they do not consider the relative proportion of training data in each class. To address this issue in this project the network was trained using Focal Loss as the loss function as it can dynamically scale Cross-Entropy. Focal Loss is designed to down-weight easy training instances and accurately classify harder, misclassified instances (Lin et al., 2018). 

\[
L_{\text{focal}}(p_t) = -\alpha_t (1 - p_t)^\gamma \log(p_t)
\]
\textbf{Equation 3:} Focal Loss \( L_{\text{focal}}(p_t) \), where \( p_t \) is the model’s predicted class probability. The parameter \( \alpha \) controls the relative importance of each class during training and \( \gamma \) controls the relative importance of harder, misclassified instances during training (Lin et al., 2018).

\section{Implementation}
All experiments were performed in a Google Colab (‘Google Colab’, n.d.) environment equipped with an Intel Xeon CPU @ 2.00GHz and NVIDIA Tesla T4 GPU with 16 GB of VRAM, and 12.67 GB of system RAM. The operating system was Ubuntu 22.04.3 LTS. The model was developed using Python 3.10.12 and PyTorch 2.3.1 (with CUDA 12.1 support) (Paszke et al., 2019). Scikit-learn 1.3.2 (Pedregosa et al., 2011), and Pandas 2.1.4 were utilized for data processing and evaluation. 

\subsection{Network Initialisation}
Weights in all layers are set at the start of training by Kaiming initialization (He et al., 2015), except for the GCN, which is initialized with a small constant value to simulate equal node treatment during initial graph structure learning (Zhang et al., 2019).  Kaiming initialization maintains activation variance and is especially useful for ReLU layers (He et al., 2015). All experiments were performed with a predefined manual seed for the random number generator to ensure reproducibility (Picard, 2023). To ensure consistency, hidden and cell states of the LSTM are initialized to zeros at the start of each sequence.

\subsection{Network Optimisation}
Network model training used the Adam optimizer to learn weights. The Adam optimizer was chosen for its adaptive learning rate to enable efficient model convergence (Kingma and Ba, 2017). A scheduler adjusted the learning rate when validation loss plateaued to ensure continuous model improvement without overshooting (C. Kim et al., 2021). 
Regularization techniques were employed to prevent the model from becoming overly complex. The final loss function, shown in Equation 4, was comprised of Focal Loss and weight decay, a form of L2 regularization, to discourage overfitting (Cortes et al., 2012). Additionally, gradient clipping was utilized to prevent exploding gradients by scaling the gradients to a maximum norm, which ensures stable and controlled updates during training (Zhang et al., 2020).

\[
L_{\text{total}} = L_{\text{focal}} + \frac{\lambda}{2} \sum_i w_i^2
\]
\textbf{Equation 4:} \( L_2 \) regularisation penalizes large weights by adding a term proportional to the sum of the squares of the weights to the loss function, where \( \lambda \) is the weight decay coefficient.
The dataset was split using stratified sampling (80\% training, 10\% validation, 10\% test) to preserve class distribution and ensure reliable evaluation. Training was performed for a total of 100 epochs. Early stopping monitored validation loss, and halting training if no improvement was seen over 10 epochs to prevent overfitting (Prechelt, 2012).

\subsection{Data Augmentation}
Data augmentation, including horizontal flipping of keypoints, was used to increase the training dataset size and improve model generalization. Due to class imbalance, bootstrapping oversampled the minority class for training and validation to create class-balanced batches and prevent model bias. The test set remained unaltered to reflect realistic distributions to ensure accurate and reliable classification outcomes.\\

\subsection{Hyperparameter Optimisation}
\begin{table}
\caption{Network hyperparameters tuned based on classification accuracy for the training dataset. Hyperparameters selected in the final model are in bold.}
\centering
\label{table:hyperparameters}
\begin{tabular}{l c c}
\toprule
\textbf{Hyperparameter} & \textbf{Description} & \textbf{Values} \\
\midrule
n\_part & Number of features per part in higher dimensional space & 32, 64, 128 \\
n\_graph\_out & Number of output features from the GCN layer & 16 \\
n\_rnn & Hidden size of the LSTM & 8, 32, 128 \\
n\_rnn\_out & Number of output features from the final BiRNN layer & 15 \\
\bottomrule
\end{tabular}
\end{table}

Hyperparameters listed in Table 3 related to the complexity of features extracted by the network were tuned to balance network complexity and model performance against overfitting to the training set (LaValle et al., 2004). A grid search was used to systematically explore and identify the optimal network settings. After selecting the best hyperparameters (Table 4), model performance was evaluated on the test set to confirm its generalization ability. Dropout layers were added to prevent overfitting by randomly dropping neurons during training, encouraging the network to learn more robust features (Srivastava et al., 2014).

\begin{table}
\caption{Hyperparameter Values}
\centering
\label{table:hyperparameters_values}
\begin{tabular}{l c}
\toprule
\textbf{Hyperparameter} & \textbf{Values} \\
\midrule
Alpha & \textbf{0.1}, 0.5, 0.9 \\
Gamma & \textbf{0}, 1, 2, 3, 5 \\
Learning rate & 0.001, \textbf{0.0001} \\
Weight decay & 0.001, \textbf{0.0001}, 0.00001 \\
\bottomrule
\end{tabular}
\end{table}

\section{Experimental Design}
\subsection{Sampling Frame Rate}
Extracted joint positions were downsampled videos from their original frame rate to a target frame rate by retaining a calculated number of time points within a "block" and skipping the rest. For example, downsampling from 25 fps to 10 fps involved skipping 15 frames and retaining 10 per block. The model was trained from scratch from each of the sampling frame rates. Model performance was compared as described in Section 2.4.2.\\

\subsection{Performance Metrics}
Model performance was performed quantitatively using the following metrics accuracy, sensitivity, and specificity (Düntsch and Gediga, 2019). Accuracy defined in Equation 5 is an overall performance metric. Sensitivity and specificity evaluate the model on its ability to correctly identify positive and negative cases respectively and are particularly important to verify model performance for imbalanced datasets (Baratloo et al., 2015).\\

\[
\text{Accuracy} = \frac{\text{TP} + \text{TN}}{\text{TP} + \text{TN} + \text{FP} + \text{FN}}
\]
\textbf{Equation 5:} Calculates the true positive and true negative in all evaluated cases. TP = True positive, TN = True negative, FP = False positive, FN = False negative.

\[
\text{Sensitivity} = \frac{\text{TP}}{\text{TP} + \text{FN}}
\]
\textbf{Equation 6:} Calculates the correct identification of the positive class out of all instances of the positive class.

\[
\text{Specificity} = \frac{\text{TN}}{\text{TN} + \text{FP}}
\]
\textbf{Equation 7:} Calculates the correct identification of the negative class out of all instances of the negative class.

\[
\text{F1 Score} = 2 \times \frac{\text{Precision} \times \text{Recall}}{\text{Precision} + \text{Recall}}
\]

\[
\text{Precision} = \frac{\text{TP}}{\text{TP} + \text{FP}}
\]

\[
\text{Recall} = \frac{\text{TP}}{\text{TP} + \text{FN}}
\]
\textbf{Equation 8:} The F1 score is the harmonic mean of precision and recall, providing a score that balances the trade-off between the two. Precision measures the accuracy of the positive predictions, while recall measures the ability of the model to find all relevant positive instances.

F1-score was also calculated, which can help to evaluate model performance on imbalanced datasets (Christen et al., 2024).\\

\subsection{Cross-validation and STatistical Ananlysis}
Cross-validation was employed to evaluate model performance (Refaeilzadeh et al., 2009). The dataset was split into five stratified folds to ensure that each fold maintained the same class distribution. For each iteration, the model was trained on four folds and validated on the remaining fold. This process was repeated for all folds, allowing the model to be evaluated on different subsets of the data. Performance metrics were averaged across all folds to provide a robust estimate of the model's generalization ability. This method helps to minimize overfitting and ensures that the model's performance is not reliant on a single data split.
After cross-validation, the model was fine-tuned using a validation set which was not involved in the cross-validation process to ensure that the model was optimally tuned before its final evaluation. The final model was then tested on an independent test set, which had been withheld during training and validation to provide an unbiased assessment of the model's performance on unseen data.
Bootstrap resampling is used to estimate confidence intervals. 1,000 bootstrap samples are created from the metrics computed by the folds. The average and 95\% confidence interval, corresponding to the 2.5th and 97.5th percentile, of the bootstrap samples are computed. This provides a range within which the true model performance is expected to fall with 95\% confidence (Efron and Tibshirani, 1986).
The p-values are calculated to compare the observed performance metrics against random guessing (50\% accuracy) (Dahiru, 2011). A one-sample t-test is computed to determine if each metric is statistically significant (p < 0.05) from the baseline.

\section{Data Visualisation}
Input data and associated temporal and spatial attention were displayed as described in the following sections to help validate the model in a transparent and clinically meaningful way.\\

\subsection{Input Data}
Visualising skeletal data helps to assess input data and model integrity as well as helping to improve model interpretability. Keypoints, represented by x, y coordinate pairs, were plotted on 0 to 1 axis and are colour coded for clarity. Neighbouring points had lines drawn between them to generate a wire frame figure. Note in this method, some figures appear distorted as aspect ratio is not preserved.\\

\subsection{Spatial Attention}
Attention scores from the model corresponding to the spatial body part outputs were used to colour keypoints (represented as described in Section 2.6.1) and depict which body parts are most informative toward the final decision.\\

The original 5x5 attention matrix, representing interactions between body parts (arms, legs and torso), was aggregated by summing elements corresponding to one body part excluding self-attention (diagonal values).\\

To improve interpretability, Gaussian smoothing was applied to the attention scores. This helps reduce jitter in colour changes and enhance meaningful patterns in the attention scores.  Temporal smoothing with a specified window size was also implemented to minimize flickering caused by rapid changes in attention across consecutive frames.\\

Attention scores were normalized with the range being carefully selected to capture subtle variations in score while disregarding outliers. The attention scores where then related to a corresponding colour in the Viridis colormap to assign to the body part. The Viridis colormap due to its wide palette, perceptual uniformity, and accessibility for colour-blind viewers (Nuñez et al., 2018; ‘matplotlib colormaps’, n.d.)\\

\subsection{Temporal Attention}
A temporal attention score bar is displayed at the bottom of each frame, using the Viridis colormap. The temporal attention score is derived from the 75x75 attention matrix computed for each video, with minimal smoothing, to preserve detail across frames. This bar shows the attention score for each frame in the video relative to the current frame and highlights when in the video the model is focused to make the final class prediction.\\

\chapter{Experimental Results}

\ifpdf
    \graphicspath{{Chapter3/Figs/Raster/}{Chapter3/Figs/PDF/}{Chapter3/Figs/}}
\else
    \graphicspath{{Chapter3/Figs/Vector/}{Chapter3/Figs/}}
\fi

\section{Model Performance}
As described in Section 2.4.2, the model was trained, and its performance was validated for different frame rates. Table 5 presents the resulting batch sizes, noting that the lowest frame rates did not use all videos due to insufficient frames to form a complete spatiotemporal matrix.\\
\begin{table}
\caption{Batch sizes for each evaluated frame rate. Each input spatiotemporal matrix contained 50 keypoint coordinates over 75 frames.}
\centering
\label{table:frame_rate}
\begin{tabular}{c c c c}
\toprule
\textbf{Frame Rate (fps)} & \textbf{Patients Skipped} & \textbf{Training Batch Size} & \textbf{Test Batch Size} \\
\midrule
5  & 20 & 74  & 26  \\
10 & 4  & 196 & 58  \\
15 & 0  & 318 & 88  \\
20 & 0  & 444 & 120 \\
25 & 0  & 558 & 148 \\
\bottomrule
\end{tabular}
\end{table}

      \begin{figure}[h!]
    \centering
    \includegraphics[width=0.7\textwidth]{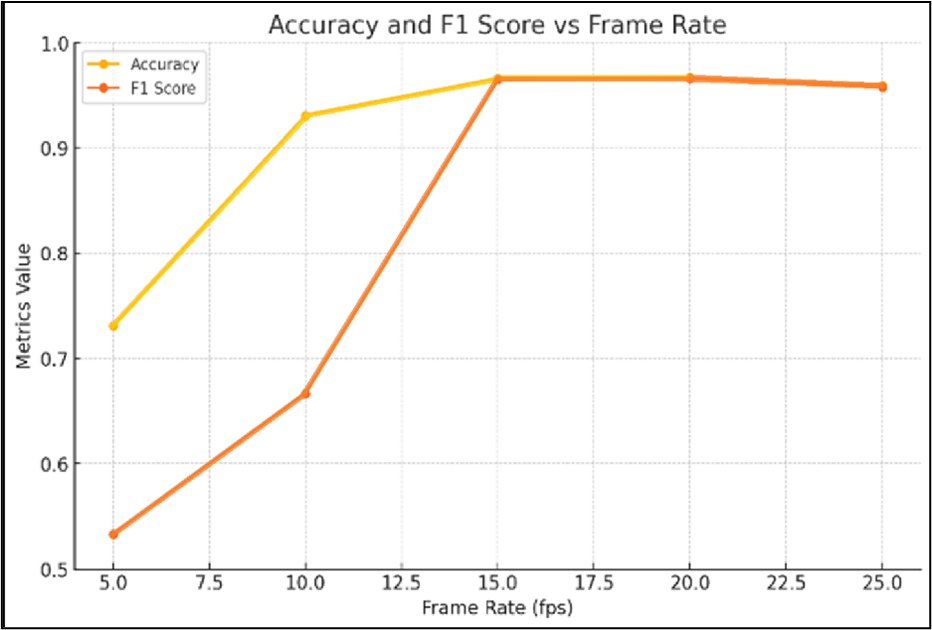}
    \caption{Graph showing the accuracy and F1-score computed for the test set for the specified frame rates.}
    \label{fig:image-label}
\end{figure}

      \begin{figure}[h!]
    \centering
    \includegraphics[width=0.7\textwidth]{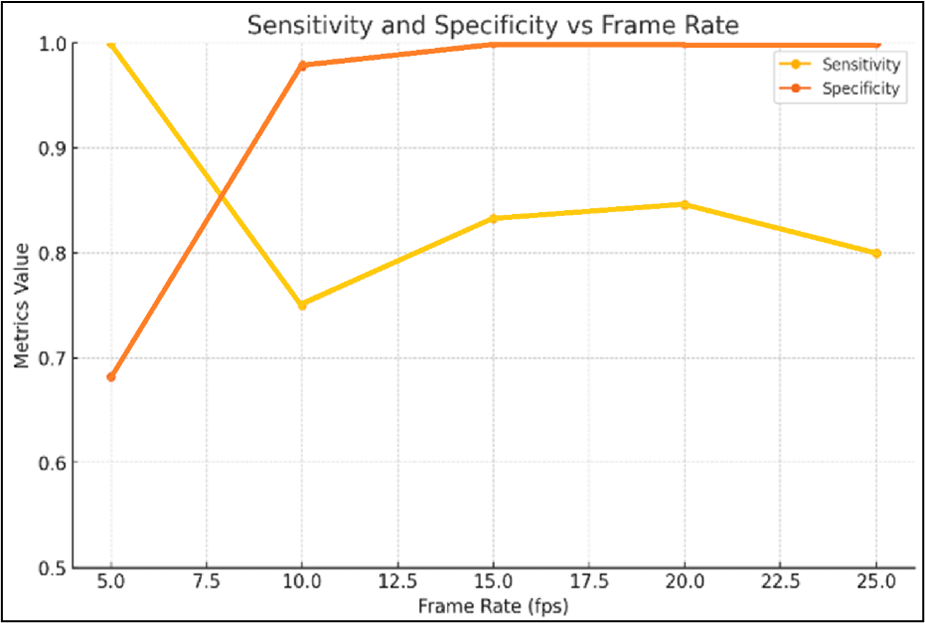}
    \caption{Graph showing the sensitivity and specificity computed for the test set for the specified frame rates.}
    \label{fig:image-label}
\end{figure}

15 fps provided the best performance for accuracy, F1-score and specificity. It must be noted that while 5 fps gave the best sensitivity, this was at the cost of specificity.
The inputs corresponding to 15 fps were used to train the model using cross-validation and perform statistical analysis. Table 6 reports the performance metrics and computed on the test set. All metrics were in the range of 80-87\% and were determined to be statistically significantly better than random guessing. 

\begin{table}
\caption{Cross-Validation Metrics with Confidence Intervals and p-values}
\centering
\label{table:cross_validation_metrics}
\begin{tabular}{l c c c}
\toprule
\textbf{Metric} & \textbf{Value} & \textbf{95\% CI} & \textbf{p-value} \\
\midrule
Accuracy     & 0.850 & (0.797, 0.906) & 0.0004 \\
F1 Score     & 0.811 & (0.743, 0.875) & 0.0013 \\
Sensitivity  & 0.810 & (0.691, 0.919) & 0.0135 \\
Specificity  & 0.878 & (0.774, 0.950) & 0.0020 \\
\bottomrule
\end{tabular}
\end{table}

      \begin{figure}[h!]
    \centering
    \includegraphics[width=0.7\textwidth]{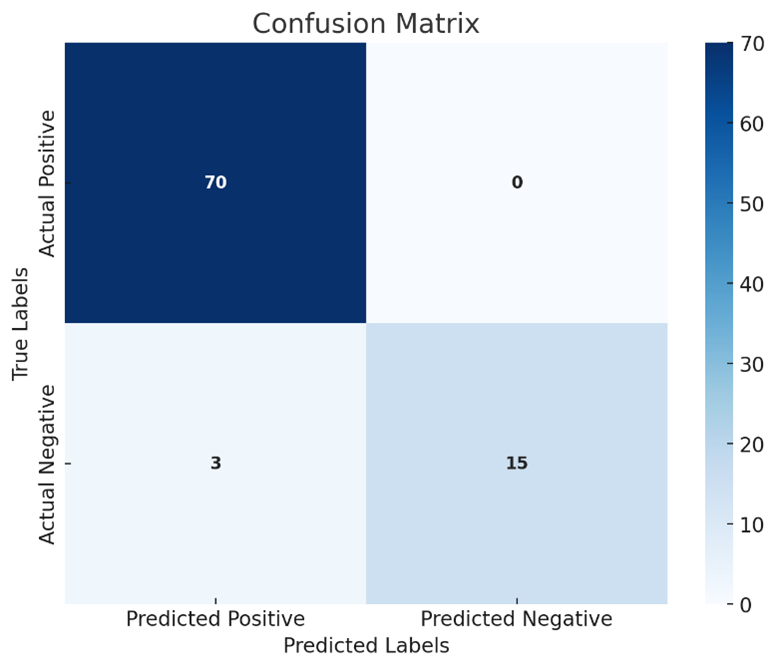}
    \caption{Confusion matrix computed for the test set of the best model at 15 fps.}
    \label{fig:image-label}
\end{figure}

To better understand the distribution of false positives and false negatives, a confusion matrix was generated for the test sets, with most of the misclassified videos occurring when dystonia-predominant HMDs (minority class) were mistaken for chorea-predominant HMDs (majority class). This is likely due to the model not having a large enough training cohort of the minority class to appropriately learn the distribution of these samples.\\

\section{Attention Mapping}
Attention maps can help to understand which body parts or frames were influential for the final classifications. The legend on the right side of each frame shows the relative importance, with brighter colours indicating higher importance (top of legend) and darker colours indicating lower importance (bottom of legend).\\

Currently, spatial and temporal attention must be considered separately. The colours of the body parts (spatial attention) help identify which parts contributed most to the final prediction. The bar at the bottom of the screen (temporal attention) highlights which point in time was most influential in classifying the videos. Brighter segments of the bar correspond to frames that were more significant in the model’s decision-making process. For example, if the bar is generally brighter when paused at a particular frame, it indicates that the model considered that frame on average more important than other frames in the video.\\

      \begin{figure}[h!]
    \centering
    \includegraphics[width=0.7\textwidth]{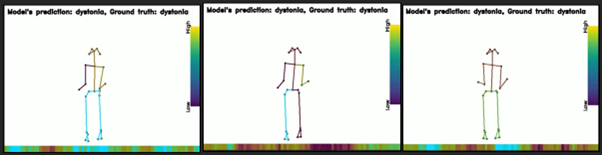}
    \caption{Frames where the input was correctly identified as dystonia. Figures a (left) and b (middle) highlight attention focused on abnormal posturing of the arm while Figure c (right) highlights attention focused on jerky, involuntary movements in the legs.}
    \label{fig:image-label}
\end{figure}

      \begin{figure}[h!]
    \centering
    \includegraphics[width=0.7\textwidth]{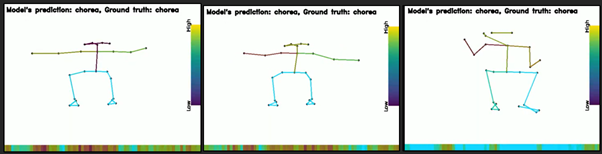}
    \caption{Frames where the input was correctly identified as chorea. Figures a (left) and b (middle) highlight attention focused on spread-out posturing in multiple parts. Figure c (right) shows diffused spatial attention due to many parts moving at once, characteristic of involuntary choreiform movements.}
    \label{fig:image-label}
\end{figure}

Figures 1-6 show frames of the correct identification of the movement disorders. Out of 45 patients identified in the training set, 20 patients had all their respective video clips correctly identified, 17 were half or more of their respective video clips correctly identified and 7 patients were under half of their respective video clips correctly identified.\\

      \begin{figure}[h!]
    \centering
    \includegraphics[width=0.7\textwidth]{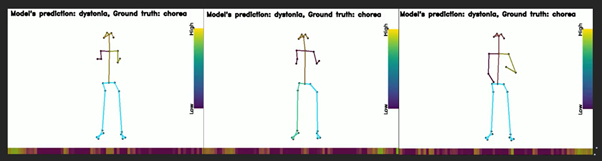}
    \caption{Frames where the input was misclassified as chorea. These frames generally have very low temporal attention scores. Additionally, spatial attention shifts between body parts in consecutive frames likely due to small perceived motions, as in dystonia. The model may have confused the patient’s posturing with that seen in dystonia.}
    \label{fig:image-label}
\end{figure}

      \begin{figure}[h!]
    \centering
    \includegraphics[width=0.7\textwidth]{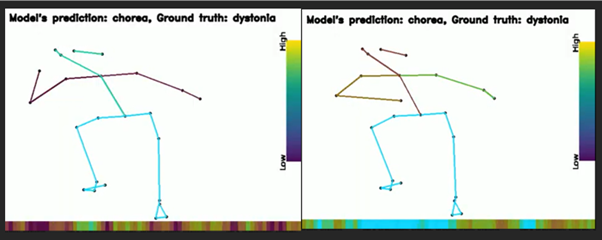}
    \caption{Frames where the input was misclassified as dystonia. Spatial attention is primarily focused on the arms. The model likely detects large amplitude movements of arms to classify these examples as chorea.
}
    \label{fig:image-label}
\end{figure}
\chapter{Discussion}

\ifpdf
    \graphicspath{{Chapter4/Figs/Raster/}{Chapter4/Figs/PDF/}{Chapter4/Figs/}}
\else
    \graphicspath{{Chapter4/Figs/Vector/}{Chapter4/Figs/}}
\fi

The primary objective of this study was to differentiate between dystonia and chorea using video data. The findings demonstrate that at 15 fps the model was able to classify samples with 85\% accuracy. All performance metrics evaluated were between 80-90\%. The qualitative results were complemented by qualitative attention maps that indicated that the model was using clinically relevant features to make decisions in most cases, hence achieving the aim of recognising the movement patterns for dystonia and chorea.\\

\section{Quantitative Analysis}
To understand the trends observed across different frame rates for the evaluation metrics, the tuning of the loss function must be considered. The value $\alpha$ =0.1  was found to achieve the best accuracy using a grid search, which means that the focal loss function significantly downweights the contribution of the majority class. This adjustment combined with bootstrapping decreased the tendency to overfit to the positive class, which consequently affects the observed sensitivity measurements.\\

The temporal characteristics of dystonic and choreiform movements also contribute to the observations. These movements unfold over a longer timeframe compared to conditions such as tremors (rapid, oscillatory movement). This temporal nature suggests that downsampling the frame rate potentially enhances the ability of the model to capture relevant movement information for these specific conditions but may not be the best frame rate for other conditions. 
Excessive downsampling (5 and 10 fps) leads to significant information loss, impacting the ability of the model to make accurate predictions. While the metrics at 15 fps onwards exhibit minimal differences, the model's overall performance appears to peak at 15 fps. At higher frame rates (25 fps), the model shows diminishing returns in sensitivity. This suggests that the additional frame information at 25 fps does not contribute significantly to the performance of the model.\\

The results from cross-validation indicated strong overall performance of the model across performance metrics. The model achieved a reasonable accuracy, sensitivity, specificity and F1-score (all >80\%). The 95\% confidence interval of all metrics forms a relatively high range (around 10\%), which indicates model variation between folds, likely due to the small dataset size used in this study. All p-values were statistically significant (p < 0.05), which indicates a statistically significant improvement over the random guessing. In this study, a more appropriate baseline was not available as this is the first study to classify dystonia and chorea in paediatric patients using video.\\

Sensitivity was more variable compared to the other metrics, due in part to the class imbalance resulting in the model having more difficulty in correctly classifying the minority class (chorea) compared to the majority class (dystonia).\\

The confusion matrix reveals that the test set consisted of more positive class samples, despite the overall dataset having more negative samples. This distribution reflects a semi-randomized distribution of classes for samples due to how patient video data are divided to form model inputs. The confusion matrix also indicated that the model produces more false positives than false negatives in its test set predictions. This is likely due to the weights of the loss function down-weighted incorrect negative predictions (i.e. false positives) rather than incorrect positive predictions (i.e. false negatives).\\

In the context of distinguishing between dystonia and chorea (as described in Section 1.2.3), there is a gap in research on automated methods for this differentiation. Studies, which are listed in Table 7, have primarily focused on distinguishing healthy individuals versus individuals with dystonia. It is unclear whether this task is inherently easier or more challenging than differentiating between dystonia and chorea. However, the task examined in this project more closely mirrors clinical practice, where healthy children are referred for examination only in rare cases.\\

      \begin{figure}[h!]
    \centering
    \includegraphics[width=0.7\textwidth]{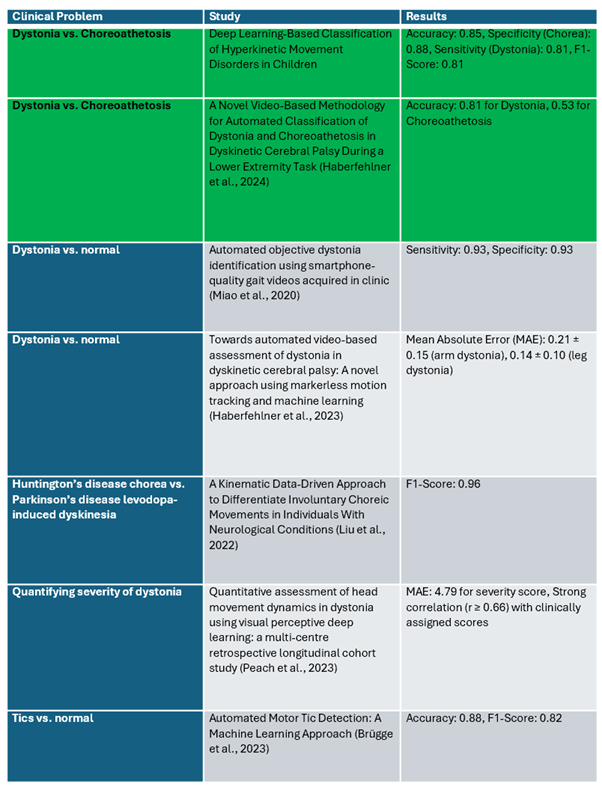}
    \caption{This summarises state-of-the-art machine learning and deep learning approaches to classify and analyse movement disorders.
}
    \label{fig:image-label}
\end{figure}

The highlighted rows in Table 4.1 compare studies distinguishing between dystonia and chorea. The first row summarizes the results of this study, and the second row summarises a study by Haberfehlner et al., 2024. Both studies address the same clinical problem, using similar metrics and cross-validation model training. Overall, this study achieves a higher accuracy (0.85) and F1-score (0.81) compared to the second study (accuracy of 0.81 for dystonia and 0.53 for choreoathetosis). The superior performance may be due to using advanced deep learning techniques and a larger dataset (50 vs. 33 patients), which helped this study’s model learn appropriate features to distinguish between these two patient populations.
While some studies focus on MAE of joint positions, an important metric for continuous monitoring post-diagnosis, the focus of this study is on classification accuracy, so there was no evaluation of how well OpenPose tracked individual joints in videos.\\

Table 4.1 also lists the various approaches used to evaluate HMDs in patients, kinematic-driven approaches achieving either significantly better results (F1-score of 0.96) or similar results in other related conditions (accuracy of 0.88 in tic disorders). Due to the flexibility of the model presented in this study, there are plans to evaluate this model, with minor modifications, to classify other HMDs in future trials.\\

\section{Qualitative Analysis}

As explained in Section 3.2, attention maps provide insight into the model predictions across body parts and video frames. Both spatial and temporal attention are critical to understanding the final model predictions. However, the interpretation of spatial attention is more intuitive as it can highlight the body parts involved in movements. Therefore, most of the qualitative analysis presented here is focused on spatial attention even though temporal attention is equally important.\\

When the model generates correct predictions, it effectively detects involuntary leg and torso movements, even though the task evaluated in this study was to keep arms outstretched. Typically, legs receive neutral attention (cyan) but shift to higher attention (green) when noticeable involuntary movements like jerks or kicks are present in the frame. Arms receive low attention (purple) when still, but attention increases (green) during stiff or jerky movements. The model rarely assigns high attention (yellow) to any single body part, reflecting the complexity of movement disorders, as there is often no single tell-tale sign and subtle leg movements may be more important for the classification than the elevation of the arms. 
The behaviour of the model is consistent with clinical decision-making (see Section 1.1.1) where it was noted that voluntary movements can sometimes exacerbate or, conversely, temporarily alleviate involuntary movements.\\

When evaluating videos where dystonia is correctly classified, the model is focused on sustained arm postures and small amplitude movements. In contrast, correct predictions of chorea are characterized by diffused attention scores across multiple body parts and reflecting irregular, large-amplitude movements. 
Notably, spatial and temporal attention do not always align. A clinically significant movement may result in higher (green) spatial attention scores, while the corresponding temporal attention remains low (purple). As mentioned in Section 2.6.3, this disparity could be due to the diffused nature of temporal scores over the attention matrix. Furthermore, because the video processing stage involved manually editing videos to include only "relevant" frames, this may have biased the temporal attention as all frames include information relevant to the task individuals were asked to perform.\\

One important sample is a video containing a "finger-to-nose" task, that was inadvertently included in the analysis. In this task, the patient alternates bringing their hand to their nose to test coordination, which is often impaired in HMDs. The model misclassified the sample as chorea due to the large amplitude movements observed in the arms. The temporal attention bar, mostly green, indicated a high focus on these large-amplitude movements at the start and end phases of the task, contrasted with the more diffused attention observed in other samples. The spatial attention highlighted arm movements as being important.\\

      \begin{figure}[h!]
    \centering
    \includegraphics[width=0.7\textwidth]{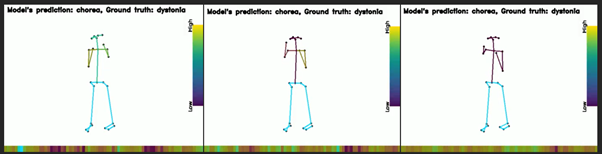}
    \caption{The accidental inclusion of the finger-to-nose task helps gain more insight into the model’s decision-making process.}
    \label{fig:image-label}
\end{figure}

      \begin{figure}[h!]
    \centering
    \includegraphics[width=0.7\textwidth]{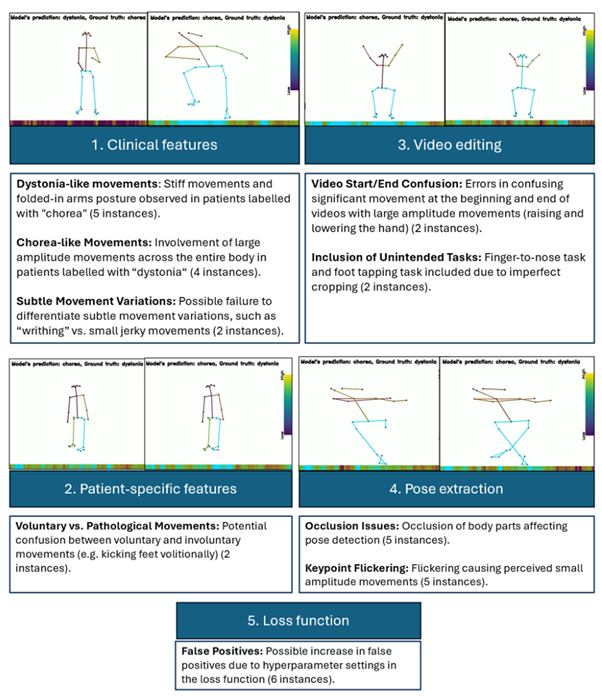}
    \caption{Potential reasons for sample misclassification. Multiple factors may contribute to the same sample being incorrectly classified.}
    \label{fig:image-label}
\end{figure}

Figure 4.3 summarizes reasons for the misclassification of samples based on a thorough review of the attention map for these samples. For samples in Box 1, the model may have confused features associated with dystonic or choreiform movements. As patients exhibit a spectrum of movements, some misclassifications involve mixed presentations with subtle differences in the final diagnosis that are hard to distinguish even for clinicians.\\

Another significant type of mistake by the model was to misinterpret normal voluntary actions (Box 2) such as moving the feet. The model may have identified such actions as pathological, large-amplitude involuntary movements. As videos are from a pediatric population, it is unrealistic to expect patients to remain motionless throughout the video recordings aside from the specified tasks. The model may require additional training examples with spurious movements unrelated to the diagnosis to prevent the model from being influenced by the presence of such movements.\\

Frames corresponding to the beginning and end of videos also have more unreliable predictions, therefore the model may interpret early movements as pathological. For instance, in one example (Box 3), the individual is slowly lowering their arms throughout the video, possibly due to fatigue. This gradual movement may have been misinterpreted as a large amplitude movement, leading to an incorrect classification as chorea.\\

Occlusion of an individual’s legs (Box 4) or shifts in keypoints across frames, due to challenges in keypoint detection also lead to erroneous classifications. These changes in keypoint coordinates may have been perceived by the model as small, jerky movements, resulting in videos being misclassified as dystonia.\\

Finally, cases where chorea was misclassified as dystonia (Box 5) are most likely due to the tuning of the loss function (see Section 4.1) causing the model to tend to default to the positive class. 
This qualitative analysis shows the challenges of accurately classifying HMDs from video data and the need for further refinement of the dataset preprocessing, the model parameters, and the training dataset to enhance accuracy.\\

\section{Real-World Impact}
This study presents one of the first deep learning models specifically designed to classify dystonia from chorea using video recordings. The model presented achieves similar or higher performance metrics compared to other models in the literature. However, an accuracy of 85\% is still lower than would be clinically acceptable.\\

This study also introduces attention maps of the model to enhance the interpretability of the model and provide clinicians with a unique opportunity to reflect on and compare their diagnostic methods against the insights generated by the model. However, the interpretation of attention maps, especially for the temporal attention map, is not straightforward and future work must ensure the relative importance of different pieces of information is clear to clinicians without deep learning expertise.\\

The model may offer a new perspective on patient evaluation, potentially helping to resolve common disagreements among clinicians and reduce the subjectivity inherent in the diagnostic process. By prompting discussions on the validity of the model’s diagnostic pathways, the network could contribute to a consensus on what specific movements constitute a clinical feature, thereby refining diagnostic protocols.\\

Misclassifications between healthy individuals and those with severe conditions could lead to unnecessary treatments, expose patients to adverse drug effects, or delay critical interventions. The applicability of the current model is limited by its training on only severely ill patients, making it unsuitable for use in a general population without further training and validation. The model must undergo rigorous validation with a larger and more diverse patient population, and scrutiny by multiple consultant neurologists and machine learning experts before deployment in a real-world clinical setting.\\

The differentiation between dystonia and chorea also carries important treatment implications. While some treatments overlap (eg. deep brain stimulation), others are directly contrasting, such as dopaminergic agents versus dopamine-depleting agents. Misdiagnosis in these cases could exacerbate the patient’s condition, which emphasizes the importance of using any classification model as a complementary tool alongside clinical judgment.\\

Deep learning classification models, such as the one developed in this study, hold significant promise as a training tool for clinicians and medical trainees, particularly given the rarity with which these HMDs are encountered outside of speciality centres. Senior clinicians could create their attention maps to highlight body parts and frames deemed critical in making a diagnosis, and then use this data to help trainees understand what clinical features contribute to a decision.\\

This study may encourage the development of deep learning techniques for other clinical applications, where video helps to capture motion pathology such as epilepsy/seizures or stroke assessment.  
The model, developed using free, open-source software and without requiring significant GPU resources, contributes to the sustainability and cost-effectiveness of the study. However, the use of a small dataset influences both of these factors. While the model cannot currently train or predict in real-time, it can produce results within the timeframe of a clinical consultation.\\

This study reported the optimal frame rate which captured the relevant temporal dynamics for diagnosis of chorea and dystonia. Future trials could benefit from recording or downsampling videos to 15 fps before performing pose extraction to optimise processing times and storage space requirements. However, care must be taken to extrapolate these findings to other HMDs that were not considered in this study.\\

In conclusion, while this study marks an advancement in the use of deep learning for classifying dystonia and chorea, but requires further refinement, and validation on larger and more diverse datasets, and must meet ethical and regulatory requirements before clinical adoption. This line of research could foster further collaboration between machine learning experts and consultant neurologists, creating new research opportunities.\\

\section{Limitations}
The study used videos from a specific patient population collected at a single tertiary hospital, which mainly examines severely ill patients. This introduces a selection bias in the training datasets, limiting the generalisation ability of the model and the key findings of this study. Additional factors that constrain generalizability include limited task variety, a small dataset size, class imbalance within the dataset, and explicitly removing videos of poor quality from consideration at the outset of the study. Classification of all patients into chorea-dominant and dystonia-dominant were performed by one consultant neurologist. Furthermore, cropping videos and restricting the start and end of the video to only the task also limits the generalisation ability of the model when presented with either non-task or different task motions.\\

The use of advanced deep learning layers, including bidirectional LSTMs, multi-head attention layers and the GCN, demand significant computational resources. Moreover, the bootstrapping and data augmentation used in this study may increase the risk of model overfitting, particularly with the small, imbalanced dataset used in this study. Model performance is highly dependent on accurate hyperparameter tuning using grid search, which is computationally expensive. Additionally, while techniques like bootstrapping and Focal Loss address class imbalance, they introduce new biases (as seen with the positive class in this experiment) requiring careful adjustment.\\

Although visualization through attention maps aimed to highlight the specific body parts and frames contributing to the model prediction, understanding and interpreting the clinical significance is difficult for non-technical experts.
By tackling these limitations, the model's strength and clinical significance could be significantly improved, making it more viable for implementation in real-world situations.\\

\section{Equality, Diversity and Inclusion}

      \begin{figure}[h!]
    \centering
    \includegraphics[width=0.7\textwidth]{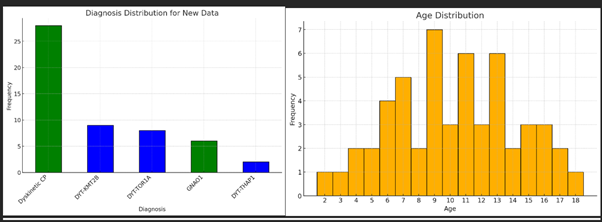}
    \caption{Distribution of specific causes of dystonia (blue) and chorea (green) in the dataset (left). Age distribution of patients in the dataset (right).}
    \label{fig:image-label}
\end{figure}
The age, gender and specific underlying diagnosis for each patient were evaluated. The demographic information is crucial as it influences the type of clinical presentations assessed by the model. Additionally, the presentation of clinical features could have subtle differences in patients of different ages and genders. The model was trained on a pediatric population, aged 2 to 18, with no obvious skew and with equal representation of male and female patients (25 each). This ensures the model's applicability across different ages and genders. Currently, the dataset includes more cases of dyskinetic cerebral palsy, which reflects its higher prevalence.  Therefore, future training should aim to incorporate more genetic diagnoses, especially those associated with dystonia, to enhance the model's generalisability. In this study, the ethnicity of the patients considered in the dataset was not collected or analysed. However, this will be considered in future work as bias in deep learning model performance related to ethnicity has been reported for other models (Buolamwini and Gebru, 2018).\\

\section{Future Work}
Future research in the long term will focus on acquiring larger, more diverse datasets that include healthy control subjects to enhance model performance and generalisation ability. Potentially, publicly available datasets of normal actions could be used for training, where it is matched to similar actions used for diagnosis. Additionally, incorporating different or more tasks from the clinical examination could provide a more comprehensive understanding of patient movement patterns and improve overall diagnostic accuracy.\\

In the short term, additional data augmentation techniques should be evaluated to generate more synthetic data. Transfer learning could potentially reduce the need for a large training dataset by learning model weights from a related dataset, such as action recognition datasets, and then adapting these weights to the specific diagnostic task. The relative contribution of the temporal and spatial network to overall model performance should be evaluated using an ablation study to understand if the complex network presented in this work is required to achieve high diagnostic accuracy.\\

Furthermore, clinical validation is essential to ensure the model complements existing diagnostic tools. The model could be adapted for real-time detection, which is particularly crucial in emergency settings, such as in cases involving GNAO1, but further analysis of the impact on clinical care should be considered first. Developing more intuitive methods to illustrate the decision-making process of the model would enhance the explainability, and should be primarily focused on the temporal features, as the current attention map is difficult to interpret. Finally, integrating longitudinal patient data into the model could enable tracking the severity of HMDs over time, or even evaluating how treatments impact the condition.\\

\chapter{Conclusion}

\ifpdf
    \graphicspath{{Chapter5/Figs/Raster/}{Chapter5/Figs/PDF/}{Chapter5/Figs/}}
\else
    \graphicspath{{Chapter5/Figs/Vector/}{Chapter5/Figs/}}
\fi

This study introduces a novel deep-learning model for classifying dystonia and chorea, showing significant potential to enhance the accuracy and objectivity of movement disorder diagnostics. The model outperforms traditional methods and offers valuable insights through attention mapping, which could refine and standardize diagnostic protocols.
However, several limitations must be addressed before clinical adoption. The use of limited patient data from a single institution and the lack of healthy controls highlights the need for broader data inclusion. To mitigate health and safety risks, the risk of giving incorrect diagnoses and treatments must be managed. The environmental impact of training large deep learning models and the commercial risks related to data privacy and regulatory compliance must also be considered. Future work should focus on expanding the dataset, exploring alternative and efficient deep-learning methods and improving the model’s explainability. Integrating feedback from clinicians and adapting the model to diverse tasks and settings will allow broader implementation.
In summary, while this research represents a significant advancement in deep learning-driven diagnostics for movement disorders, it emphasises the need for ongoing refinement, extensive validation, and careful risk management. This work lays a strong foundation for future developments that could transform clinical practice.

\chapter{References}

\ifpdf
    \graphicspath{{Chapter6/Figs/Raster/}{Chapter6/Figs/PDF/}{Chapter6/Figs/}}
\else
    \graphicspath{{Chapter6/Figs/Vector/}{Chapter6/Figs/}}
\fi

Albanese, A. et al. (2006) A systematic review on the diagnosis and treatment of primary (idiopathic) dystonia and dystonia plus syndromes: report of an EFNS/MDS‐ES Task Force. European Journal of Neurology. [Online] 13 (5), 433–444. [online]. Available from: https://onlinelibrary.wiley.com/doi/10.1111/j.1468-1331.2006.01537.x (Accessed 28 August 2024).\\
Anon (n.d.) AI Is Being Built on Dated, Flawed Motion-Capture Data - IEEE Spectrum [online]. Available from: https://spectrum.ieee.org/motion-capture-standards (Accessed 28 August 2024).\\
Anon (n.d.) Google Colab [online]. Available from: https://colab.research.google.com/drive/
1rPk3ohrmVclqhH7uQ7qys4oznDdAhpzF (Accessed 3 June 2024).\\
Anon (n.d.) GSTT Electronic Records Research Interface (GERRI) [online]. Available from: https://www.hra.nhs.uk/planning-and-improving-research/application-summaries/research-summaries/gstt-electronic-records-research-interface-gerri/ (Accessed 28 August 2024).\\
Anon (n.d.) matplotlib colourmaps [online]. Available from: https://bids.github.io/colormap/ (Accessed 28 August 2024).\\
Anon (n.d.) Official download of VLC media player, the best Open Source player - VideoLAN [online]. Available from: https://www.videolan.org/vlc/index.html (Accessed 28 August 2024).\\
Badheka, R. et al. (2018) Pediatric movement disorders. Neurology India. [Online] 66 (7), 59. [online]. Available from: https://journals.lww.com/10.4103/0028-3886.226447 (Accessed 28 August 2024).\\
Baratloo, A. et al. (2015) Part 1: Simple Definition and Calculation of Accuracy, Sensitivity and Specificity. Emergency (Tehran, Iran). 3 (2), 48–49.\\
Barry, M. J. et al. (1999) Reliability and responsiveness of the Barry–Albright Dystonia Scale. Developmental Medicine \& Child Neurology. [Online] 41 (6), 404–411. [online]. Available from: http://doi.wiley.com/10.1017/S0012162299000870 (Accessed 28 August 2024).\\
Brügge, N. S. et al. (2023) Automated Motor Tic Detection: A Machine Learning Approach. Movement Disorders. [Online] 38 (7), 1327–1335. [online]. Available from: https://movementdisorders.onlinelibrary.wiley.com/doi/10.1002/mds.29439 (Accessed 28 August 2024).\\
Buolamwini, J. \& Gebru, T. (2018) ‘Gender Shades: Intersectional Accuracy Disparities in Commercial Gender Classification’, in Proceedings of the 1st Conference on Fairness, Accountability and Transparency. 21 January 2018 PMLR. pp. 77–91. [online]. Available from: https://proceedings.mlr.press/v81/buolamwini18a.html (Accessed 28 August 2024).\\
Cao, Z. et al. (2019) OpenPose: Realtime Multi-Person 2D Pose Estimation using Part Affinity Fields. [online]. Available from: http://arxiv.org/abs/1812.08008 (Accessed 28 August 2024).
Chen, Y. et al. (2020) Monocular human pose estimation: A survey of deep learning-based methods. Computer Vision and Image Understanding. [Online] 192102897. [online]. Available from: https://linkinghub.elsevier.com/retrieve/pii/S1077314219301778 (Accessed 28 August 2024).\\
Christen, P. et al. (2024) A Review of the F-Measure: Its History, Properties, Criticism, and Alternatives. ACM Computing Surveys. [Online] 56 (3), 1–24. [online]. Available from: https://dl.acm.org/doi/10.1145/3606367 (Accessed 28 August 2024).\\
Chung, J.-L. et al. (2022) Comparative Analysis of Skeleton-Based Human Pose Estimation. Future Internet. [Online] 14 (12), 380. [online]. Available from: https://www.mdpi.com/
1999-5903/14/12/380 (Accessed 28 August 2024).\\
Cortes, C. et al. (2012) L2 Regularization for Learning Kernels. [online]. Available from: http://arxiv.org/abs/1205.2653 (Accessed 28 August 2024).\\
Dahiru, T. (2011) P-Value, a true test of statistical significance? a cautionary note. Annals of Ibadan Postgraduate Medicine. [Online] 6 (1), 21–26. [online]. Available from: http://www.ajol.info/index.php/aipm/article/view/64038 (Accessed 28 August 2024).\\
De Micco, F. \& Scendoni, R. (2024) Three Different Currents of Thought to Conceive Justice: Legal, and Medical Ethics Reflections. Philosophies. [Online] 9 (3), 61. [online]. Available from: https://www.mdpi.com/2409-9287/9/3/61 (Accessed 28 August 2024).
Defferrard, M. et al. (2016) ‘Convolutional neural networks on graphs with fast localized spectral filtering’, in Proceedings of the 30th International Conference on Neural Information Processing Systems. NIPS’16. 5 December 2016 Red Hook, NY, USA: Curran Associates Inc. pp. 3844–3852.\\
Duan, H. et al. (2022) Revisiting Skeleton-based Action Recognition. [online]. Available from: http://arxiv.org/abs/2104.13586 (Accessed 28 August 2024).\\
Düntsch, I. \& Gediga, G. (2019) Confusion matrices and rough set data analysis. Journal of Physics: Conference Series. [Online] 1229 (1), 012055. [online]. Available from: http://arxiv.org/abs/1902.01487 (Accessed 28 August 2024).\\
Efron, B. \& Tibshirani, R. (1986) Bootstrap Methods for Standard Errors, Confidence Intervals, and Other Measures of Statistical Accuracy. Statistical Science. [Online] 1 (1),. [online]. Available from: https://projecteuclid.org/journals/statistical-science/\\
volume-1/issue-1/Bootstrap-Methods-for-Standard-Errors-
Confidence-Intervals-and-Other-Measures/
10.1214/ss/1177013815.full (Accessed 28 August 2024).\\
Fernández-Alvarez, E. \& Nardocci (2012) Update on pediatric dystonias: etiology, epidemiology, and management. Degenerative Neurological and Neuromuscular Disease. [Online] 29. [online]. Available from: http://www.dovepress.com/update-on-pediatric-dystonias-etiology-epidemiology-and-management-peer-reviewed-article-DNND (Accessed 21 August 2024).\\
Gatto, E. M. et al. (2021) Worldwide barriers to genetic testing for movement disorders. European Journal of Neurology. [Online] 28 (6), 1901–1909. [online]. Available from: https://onlinelibrary.wiley.com/doi/10.1111/ene.14826 (Accessed 28 August 2024).\\
Gorodetsky, C. \& Fasano, A. (2022) Approach to the Treatment of Pediatric Dystonia. Dystonia. [Online] 110287. [online]. Available from: https://www.frontierspartnerships.org/journals/dystonia/articles/10.3389/dyst.2022.10287/full (Accessed 28 August 2024).\\
Haberfehlner, H. et al. (2024) A Novel Video-Based Methodology for Automated Classification of Dystonia and Choreoathetosis in Dyskinetic Cerebral Palsy During a Lower Extremity Task. Neurorehabilitation and Neural Repair. [Online] 38 (7), 479–492. [online]. Available from: https://journals.sagepub.com/doi/10.1177/15459683241257522 (Accessed 28 August 2024).\\
Haberfehlner, H. et al. (2023) Towards automated video-based assessment of dystonia in dyskinetic cerebral palsy: A novel approach using markerless motion tracking and machine learning. Frontiers in Robotics and AI. [Online] 10. [online]. Available from: https://www.frontiersin.org/articles/10.3389/frobt.2023.1108114 (Accessed 12 March 2024).\\
Hara, K. et al. (2017) Learning Spatio-Temporal Features with 3D Residual Networks for Action Recognition. [online]. Available from: https://arxiv.org/abs/1708.07632 (Accessed 28 August 2024).
He, K. et al. (2015a) Deep Residual Learning for Image Recognition. [online]. Available from: http://arxiv.org/abs/1512.03385 (Accessed 28 August 2024).\\
He, K. et al. (2015b) Delving Deep into Rectifiers: Surpassing Human-Level Performance on ImageNet Classification. [online]. Available from: http://arxiv.org/abs/1502.01852 (Accessed 28 August 2024).\\
Hochreiter, S. \& Schmidhuber, J. (1997) Long Short-Term Memory. Neural Computation. [Online] 9 (8), 1735–1780. [online]. Available from: https://direct.mit.edu/neco/article/9/8/1735-1780/6109 (Accessed 21 August 2024).\\
Ioffe, S. \& Szegedy, C. (2015) Batch Normalization: Accelerating Deep Network Training by Reducing Internal Covariate Shift. [online]. Available from: http://arxiv.org/abs/1502.03167 (Accessed 28 August 2024).\\
Jin, S. et al. (2020) Whole-Body Human Pose Estimation in the Wild. [online]. Available from: http://arxiv.org/abs/2007.11858 (Accessed 28 August 2024).\\
Joshi, R. B. D. \& Joshi, D. (2021) MoveNet: A Deep Neural Network for Joint Profile Prediction Across Variable Walking Speeds and Slopes. IEEE Transactions on Instrumentation and Measurement. [Online] 701–11. [online]. Available from: https://ieeexplore.ieee.org/
document/9406043/ (Accessed 28 August 2024).\\
Kendall, A. et al. (2016) PoseNet: A Convolutional Network for Real-Time 6-DOF Camera Relocalization. [online]. Available from: http://arxiv.org/abs/1505.07427 (Accessed 28 August 2024).\\
Khemani, B. et al. (2024) A review of graph neural networks: concepts, architectures, techniques, challenges, datasets, applications, and future directions. Journal of Big Data. [Online] 11 (1), 18. [online]. Available from: https://journalofbigdata.springeropen.com/
articles/10.1186/s40537-023-00876-4 (Accessed 28 August 2024).\\
Kim, C. et al. (2021) Automated Learning Rate Scheduler for Large-batch Training. [online]. Available from: http://arxiv.org/abs/2107.05855 (Accessed 28 August 2024).\\
Kim, W. et al. (2021) Ergonomic postural assessment using a new open-source human pose estimation technology (OpenPose). International Journal of Industrial Ergonomics. [Online] 84103164. [online]. Available from: https://linkinghub.elsevier.com/retrieve/\\
pii/S0169814121000822 (Accessed 28 August 2024).\\
Kingma, D. P. \& Ba, J. (2017) Adam: A Method for Stochastic Optimization. [online]. Available from: http://arxiv.org/abs/1412.6980 (Accessed 28 August 2024).\\
Kipf, T. N. \& Welling, M. (2017) Semi-supervised classification with Graph Convolutional Networks. [online]. Available from: http://arxiv.org/abs/1609.02907 (Accessed 21 August 2024).\\
Koller, W. C. \& Biary, N. M. (1989) Volitional control of involuntary movements. Movement Disorders. [Online] 4 (2), 153–156. [online]. Available from: https://movementdisorders.onlinelibrary.wiley.com/doi/10.1002/mds.870040207 (Accessed 28 August 2024).\\
Kruer, M. C. (2015) Pediatric Movement Disorders. Pediatrics In Review. [Online] 36 (3), 104–116. [online]. Available from: https://publications.aap.org/pediatricsinreview/
article/36/3/104/34867/Pediatric-Movement-Disorders (Accessed 28 August 2024).\\
Kuiper, M. J. et al. (2016) The Burke‐Fahn‐Marsden Dystonia Rating Scale is Age‐Dependent in Healthy Children. Movement Disorders Clinical Practice. [Online] 3 (6), 580–586. [online]. Available from: https://movementdisorders.onlinelibrary.wiley.com/\\
doi/10.1002/mdc3.12339 (Accessed 28 August 2024).\\
LaValle, S. M. et al. (2004) On the Relationship between Classical Grid Search and Probabilistic Roadmaps. The International Journal of Robotics Research. [Online] 23 (7–8), 673–692. [online]. Available from: http://journals.sagepub.com/doi/10.1177/0278364904045481 (Accessed 28 August 2024).\\
Li, G. et al. (2020) Deeper GCN: All You Need to Train Deeper GCNs. [online]. Available from: http://arxiv.org/abs/2006.07739 (Accessed 28 August 2024).\\
Lin, J.-P. (2011) The contribution of spasticity to the movement disorder of cerebral palsy using pathway analysis: does spasticity matter?: Commentaries. Developmental Medicine \& Child Neurology. [Online] 53 (1), 7–9. [online]. Available from: https://onlinelibrary.wiley.com/doi/10.1111/j.1469-8749.2010.03843.x (Accessed 28 August 2024).\\
Lin, J.-P. et al. (2014) The impact and prognosis for dystonia in childhood including dystonic cerebral palsy: a clinical and demographic tertiary cohort study. Journal of Neurology, Neurosurgery \& Psychiatry. [Online] 85 (11), 1239–1244. [online]. Available from: https://jnnp.bmj.com/lookup/doi/10.1136/jnnp-2013-307041 (Accessed 28 August 2024).\\
Lin, T.-Y. et al. (2018) Focal Loss for Dense Object Detection. [online]. Available from: http://arxiv.org/abs/1708.02002 (Accessed 28 August 2024).\\
Luo, Z. et al. (2021) Rethinking the Heatmap Regression for Bottom-up Human Pose Estimation. [online]. Available from: http://arxiv.org/abs/2012.15175 (Accessed 28 August 2024).\\
Mazzia, V. et al. (2022) Action Transformer: A self-attention model for short-time pose-based human action recognition. Pattern Recognition. [Online] 124108487. [online]. Available from: https://www.sciencedirect.com/science/article/pii/\\S0031320321006634 (Accessed 3 June 2024).\\
Méneret, A. et al. (2021) Treatable Hyperkinetic Movement Disorders Not to Be Missed. Frontiers in Neurology. [Online] 12659805. [online]. Available from: https://www.frontiersin.org/articles/10.3389/fneur.2021.659805/full (Accessed 28 August 2024).\\
Merical, B. \& Sánchez-Manso, J. C. (2024) ‘Chorea’, in StatPearls. Treasure Island (FL): StatPearls Publishing. p. [online]. Available from: http://www.ncbi.nlm.nih.gov/books/NBK430923/ (Accessed 21 August 2024).\\
Miao, H. et al. (2020) Automated objective dystonia identification using smartphone-quality gait videos acquired in the clinic. p.2020.06.09.20116954. [online]. Available from: https://www.medrxiv.org/content/10.1101/2020.06.09.20116954v1 (Accessed 3 June 2024).\\
Mink, J. W. \& Sanger, T. D. (2017) ‘Movement Disorders’, in Swaiman’s Pediatric Neurology. [Online]. Elsevier. pp. 706–717. [online]. Available from: https://linkinghub.elsevier.com/retrieve/pii/B978032337101800093X (Accessed 28 August 2024).\\
Nath, T. et al. (2019) Using DeepLabCut for 3D markerless pose estimation across species and behaviours. Nature Protocols. [Online] 14 (7), 2152–2176. [online]. Available from: https://www.nature.com/articles/s41596-019-0176-0 (Accessed 12 March 2024).\\
Nikkhah, A. et al. (2019) Hyperkinetic Movement Disorders in Children: A Brief Review. Iranian Journal of Child Neurology. 13 (2), 7–16.
Noseworthy, P. A. et al. (2020) Assessing and Mitigating Bias in Medical Artificial Intelligence: The Effects of Race and Ethnicity on a Deep Learning Model for ECG Analysis. Circulation: Arrhythmia and Electrophysiology. [Online] 13 (3), e007988. [online]. Available from: https://www.ahajournals.org/doi/10.1161/CIRCEP.119.007988 (Accessed 28 August 2024).\\
Nuñez, J. R. et al. (2018) Optimizing colourmaps with consideration for colour vision deficiency to enable accurate interpretation of scientific data Jesús Malo (ed.). PLOS ONE. [Online] 13 (7), e0199239. [online]. Available from: https://dx.plos.org/10.1371/journal.pone.0199239 (Accessed 28 August 2024).\\
Osokin, D. (2018) Real-time 2D Multi-Person Pose Estimation on CPU: Lightweight OpenPose. [online]. Available from: http://arxiv.org/abs/1811.12004 (Accessed 3 June 2024).\\
Pascanu, R. et al. (2013) On the difficulty of training Recurrent Neural Networks. [online]. Available from: http://arxiv.org/abs/1211.5063 (Accessed 28 August 2024).\\
Paszke, A. et al. (2019) PyTorch: An Imperative Style, High-Performance Deep Learning Library. [online]. Available from: http://arxiv.org/abs/1912.01703 (Accessed 28 August 2024).\\
Payne, K. et al. (2018) Cost-effectiveness analyses of genetic and genomic diagnostic tests. Nature Reviews Genetics. [Online] 19 (4), 235–246. [online]. Available from: https://www.nature.com/articles/nrg.2017.108 (Accessed 28 August 2024).\\
Peach, R. et al. (2023) Quantitative assessment of head movement dynamics in dystonia using visual perceptive deep learning: a multi-centre retrospective longitudinal cohort study. p.2023.09.11.23295260. [online]. Available from: https://www.medrxiv.org/content/10.1101/\\2023.09.11.23295260v1 (Accessed 13 March 2024).\\
Pedregosa, F. et al. (2011) Scikit-learn: Machine Learning in Python. J. Mach. Learn. Res. 12 (null), 2825–2830.\\
Pérez‐Dueñas, B. et al. (2022) The Genetic Landscape of Complex Childhood‐Onset Hyperkinetic Movement Disorders. Movement Disorders. [Online] 37 (11), 2197–2209. [online]. Available from: https://www.ncbi.nlm.nih.gov/pmc/articles/PMC9804670/ (Accessed 28 August 2024).\\
Philipp, G. et al. (2018) Gradients explode - Deep Networks are shallow - ResNet explained. [online]. Available from: https://openreview.net/forum?id=HkpYwMZRb (Accessed 28 August 2024).
Picard, D. (2023) Torch.manual\_seed(3407) is all you need: On the influence of random seeds in deep learning architectures for computer vision. [online]. Available from: http://arxiv.org/abs/2109.08203 (Accessed 28 August 2024).\\
Pietracupa, S. et al. (2015) Scales for hyperkinetic disorders: A systematic review. Journal of the Neurological Sciences. [Online] 358 (1), 9–21. [online]. Available from: https://www.sciencedirect.com/science/article/pii/\\S0022510X15020407 (Accessed 28 August 2024).\\
Prechelt, L. (2012) ‘Early Stopping — But When?’, in Grégoire Montavon et al. (eds.) Neural Networks: Tricks of the Trade. Lecture Notes in Computer Science. [Online]. Berlin, Heidelberg: Springer Berlin Heidelberg. pp. 53–67. [online]. Available from: http://link.springer.com/10.1007/978-3-642-35289-8\_5 (Accessed 28 August 2024).\\
Qin, X. et al. (2022) An efficient self-attention network for skeleton-based action recognition. Scientific Reports. [Online] 12 (1), 4111. [online]. Available from: https://www.nature.com/articles/s41598-022-08157-5 (Accessed 3 June 2024).\\
Refaeilzadeh, P. et al. (2009) ‘Cross-Validation’, in Ling Liu \& M. Tamer Özsu (eds.) Encyclopedia of Database Systems. [Online]. Boston, MA: Springer US. pp. 532–538. [online]. Available from: http://link.springer.com/10.1007/978-0-387-39940-9\_565 (Accessed 28 August 2024).\\
Sanders, A. E. et al. (2024) ‘Myoclonus’, in StatPearls. Treasure Island (FL): StatPearls Publishing. p. [online]. Available from: http://www.ncbi.nlm.nih.gov/books/NBK537015/ (Accessed 28 August 2024).\\
Selvaraju, R. R. et al. (2016) Grad-CAM: Visual Explanations from Deep Networks via Gradient-based Localization. [Online] [online]. Available from: https://arxiv.org/abs/1610.02391 (Accessed 28 August 2024).\\
Shams, R. A. et al. (2023) AI and the quest for diversity and inclusion: a systematic literature review. AI and Ethics. [Online] [online]. Available from: https://link.springer.com/10.1007/s43681-023-00362-w (Accessed 28 August 2024).\\
Shu, X. et al. (2022) Spatiotemporal Co-Attention Recurrent Neural Networks for Human-Skeleton Motion Prediction. IEEE Transactions on Pattern Analysis and Machine Intelligence. [Online] 44 (6), 3300–3315. [online]. Available from: https://ieeexplore.ieee.org/document/9321130/ (Accessed 3 June 2024).\\
Si, C. et al. (2020) Skeleton-based action recognition with hierarchical spatial reasoning and temporal stack learning network. Pattern Recognition. [Online] 107107511. [online]. Available from: https://linkinghub.elsevier.com/retrieve/pii/S0031320320303149 (Accessed 28 August 2024).\\
Soydaner, D. (2022) Attention Mechanism in Neural Networks: Where it Comes and Where it Goes. Neural Computing and Applications. [Online] 34 (16), 13371–13385. [online]. Available from: http://arxiv.org/abs/2204.13154 (Accessed 28 August 2024).\\
Srivastava, N. et al. (2014) Dropout: A Simple Way to Prevent Neural Networks from Overfitting. Journal of Machine Learning Research. 15 (56), 1929–1958. [online]. Available from: http://jmlr.org/papers/v15/srivastava14a.html (Accessed 28 August 2024).\\
Stewart, K. et al. (2017) A systematic review of scales to measure dystonia and choreoathetosis in children with dyskinetic cerebral palsy. Developmental Medicine \& Child Neurology. [Online] 59 (8), 786–795. [online]. Available from: https://onlinelibrary.wiley.com/doi/10.1111/dmcn.13452 (Accessed 28 August 2024).\\
Stouwe, A. M. M. van der et al. (2021) Next move in movement disorders (NEMO): developing a computer-aided classification tool for hyperkinetic movement disorders. BMJ Open. [Online] 11 (10), e055068. [online]. Available from: https://bmjopen.bmj.com/content/11/10/e055068 (Accessed 28 August 2024).\\
Tadic, V. et al. (2012) Dopa-Responsive Dystonia Revisited: Diagnostic Delay, Residual Signs, and Nonmotor Signs. Archives of Neurology. [Online] 69 (12), 1558. [online]. Available from: http://archneur.jamanetwork.com/article.aspx?doi=10.1001/archneurol.2012.574 (Accessed 28 August 2024).\\
Vaswani, A. et al. (2023) Attention Is All You Need. [online]. Available from: http://arxiv.org/abs/1706.03762 (Accessed 28 August 2024).\\
Washabaugh, E. P. et al. (2022) Comparing the accuracy of open-source pose estimation methods for measuring gait kinematics. Gait \& Posture. [Online] 97188–195. [online]. Available from: https://linkinghub.elsevier.com/retrieve/pii/S0966636222004738 (Accessed 28 August 2024).
Wu, F. et al. (2019) Simplifying Graph Convolutional Networks. [online]. Available from: https://arxiv.org/abs/1902.07153 (Accessed 28 August 2024).\\
Wu, Z. et al. (2021) A Comprehensive Survey on Graph Neural Networks. IEEE Transactions on Neural Networks and Learning Systems. [Online] 32 (1), 4–24. [online]. Available from: https://ieeexplore.ieee.org/document/9046288/ (Accessed 28 August 2024).\\
Xu, B. et al. (2015) Empirical Evaluation of Rectified Activations in Convolutional Network. [online]. Available from: http://arxiv.org/abs/1505.00853 (Accessed 28 August 2024).\\
Yan, S. et al. (2018) Spatial Temporal Graph Convolutional Networks for Skeleton-Based Action Recognition. [online]. Available from: http://arxiv.org/abs/1801.07455 (Accessed 28 August 2024).
Yilmaz, S. \& Mink, J. W. (2020) Treatment of Chorea in Childhood. Pediatric Neurology. [Online] 10210–19. [online]. Available from: https://linkinghub.elsevier.com/retrieve/pii/S0887899419308173 (Accessed 28 August 2024).\\
Zhang, J. et al. (2020) Why gradient clipping accelerates training: A theoretical justification for adaptivity. [online]. Available from: http://arxiv.org/abs/1905.11881 (Accessed 28 August 2024).\\
Zhang, Q. et al. (2019) Learning graph structure via graph convolutional networks. Pattern Recognition. [Online] 95308–318. [online]. Available from: https://linkinghub.elsevier.com/retrieve/pii/S0031320319302432 (Accessed 28 August 2024).\\
Zhao, Z. et al. (2023) Interpretation of Time-Series Deep Models: A Survey. [online]. Available from: http://arxiv.org/abs/2305.14582 (Accessed 28 August 2024).\\










\begin{appendices} 

\chapter{GitHub Repository} 

The OpenPose data of patients with dystonia and chorea, along with the original neural network code for classification, cross-validation, and attention visualisation, are available in the GitHub repository at the following link:

\url{https://github.com/nandika-r/hmds-network}

\end{appendices}

\printthesisindex 

\end{document}